\documentclass[conference]{IEEEtran}

\usepackage{times}
\usepackage{epsfig}
\usepackage{graphicx}
\usepackage{amsmath}
\usepackage{amssymb}
\usepackage{gensymb}
\usepackage{latexsym}
\usepackage{amsbsy}

\begin{document}

\title{FMCode: A 3D In-the-Air Finger Motion Based User Login Framework for Gesture Interface}


\author{
\IEEEauthorblockN{Duo Lu}
\IEEEauthorblockA{
%School of Computing, Informatics, \\and Decision Systems Engineering\\
Arizona State University\\
%Tempe, Arizona\\
duolu@asu.edu}
\and
\IEEEauthorblockN{Dijiang Huang}
\IEEEauthorblockA{
%School of Computing, Informatics, \\and Decision Systems Engineering\\
Arizona State University\\
%Tempe, Arizona\\
dijiang.huang@asu.edu}
}

% \author{
% \IEEEauthorblockN{Anonymous submission to IEEE Symposium on Security and Privacy 2018}
% \IEEEauthorblockA{Anonymous Institute\\ anonymous@email.com}
% }

\maketitle

\thispagestyle{plain}
\pagestyle{plain}

\begin{abstract}

Applications using gesture-based human-computer interface require a new user login method with gestures because it does not have traditional input method to type a password. However, due to various challenges, existing gesture based authentication systems are generally considered too weak to be useful in practice. In this paper, we propose a unified user login framework using 3D in-air-handwriting, called FMCode. We define new types of features critical to distinguish legitimate users from attackers and utilize Support Vector Machine (SVM) for user authentication. The features and data-driven models are specially designed to accommodate minor behavior variations that existing gesture authentication methods neglect. In addition, we use deep neural network approaches to efficiently identify the user based on his or her in-air-handwriting, which avoids expansive account database search methods employed by existing work. On a dataset collected by us with over 100 users, our prototype system achieves 0.1\% and 0.5\% best Equal Error Rate (EER) for user authentication, as well as 96.7\% and 94.3\% accuracy for user identification, using two types of gesture input devices. Compared to existing behavioral biometric systems using gesture and in-air-handwriting, our framework achieves the state-of-the-art performance. In addition, our experimental results show that FMCode is capable to defend against client side spoofing attacks, and it performs persistently in the long run. These results and discoveries pave the way to practical usage of gesture based user login over the gesture interface.

\end{abstract}


% \fi
\IEEEpeerreviewmaketitle

\section{Introduction}

Gesture interfaces are generally considered as the next generation of Human-Computer Interface (HCI) which can fundamentally change the way we interact with computer. Moreover, platforms equipped with gesture interfaces such as home entertainment consoles (\textit{e.g.,} Microsoft XBox with Kinect \cite{Kinect}), Virtual Reality (VR) headsets (\textit{e.g.,} HTC Vive \cite{HTCVive}), and wearable computers (e.g., Google Soli \cite{Soli}) are becoming popular in various application scenarios. On these platforms, user authentication and identification are basic functions for accessing resources, personalized services, and private data. For example, a user may be asked to verify the identity to unlock a wearable device similar as unlock a smartphone, or sign in a virtual site over a VR headset. However, in such an environment, it is usually inconvenient or impractical to access a keyboard or virtual keyboard on a touchscreen. Moreover, in certain scenarios such as clean room or operating theater, gesture interface is used to avoid physical touch due to high cleanliness. Therefore, research needs to answer how to login securely and conveniently with the gesture interface instead of using a keyboard or touchscreen. Current VR headsets (\textit{e.g.,} Oculus Rift and Vive) and wearable devices rely on the connected desktop computer or smartphone to complete the login procedure using either passwords or biometrics, which is inconvenient. A few standalone VR headsets (\textit{e.g.,} Microsoft Hololens \cite{HoloLens}) present a virtual keyboard in the air and ask the user to type a password. However, it is slow and user unfriendly due to the lack of key stroke feedback and a limited recognition accuracy of the key type action in-the-air. Moreover, passwords have their own drawbacks due to the trade-off between the memory difficulty and the password strength requirement. Biometrics employ the information strongly linked to the person, which cannot be revoked upon leakage and may raise privacy concerns in online login. Therefore, the usage of feature-rich behavioral biometrics such as the in-air-handwriting is preferable.

During the past decades, identity verification methods involving writing the password in the air have been studied on different input interfaces such as hand-held devices \cite{Madrid-Analysis, uWave, Greece}, cameras \cite{VSig, LeapHand, LeapPassword, KinWrite}, and touchscreens \cite{frank2013touchalytics, gong2016forgery, yang2016free}. However, the application of such a method encounters a few fundamental challenges. First, gesture input sensors have limited capabilities in capturing hand movements (\textit{e.g.,} limitation in accuracy, resolution, sampling speed, field of view, \textit{etc.}). Meanwhile the user's writing behavior has uncertainty in posture and magnitude. These facts make signal processing and feature extraction difficult. Second, the captured handwriting contains minor variations in speed and shape even for the same user writing the same content. Unlike the password that does not tolerate a single bit difference, this ambiguity in the in-air-handwriting leads to difficulty in designing matching algorithms and limited discriminating capability. Hence, existing solutions rarely achieve an acceptable authentication performance. Third, user identification requires indexing a large amount of accounts using the ambiguous, inaccurate and noisy in-air-handwriting motion signals in order to efficiently locate the desired account upon a login request, which cannot be accomplished directly by current template matching based methods. As a result, existing solutions have to search the account database exhaustively to compare the signal in the login request with the template, which is impractical for real world usage. Fourth, for data-driven methods that train a model to recognize each account and classify the handwriting, the available signal samples at registration are scarce because of usability consideration. Since we can only ask the user to write the passcode a few times to sign up), the effectiveness of model training is significantly restricted.

\begin{figure*}[!t]
\centering
\includegraphics[width=7in]{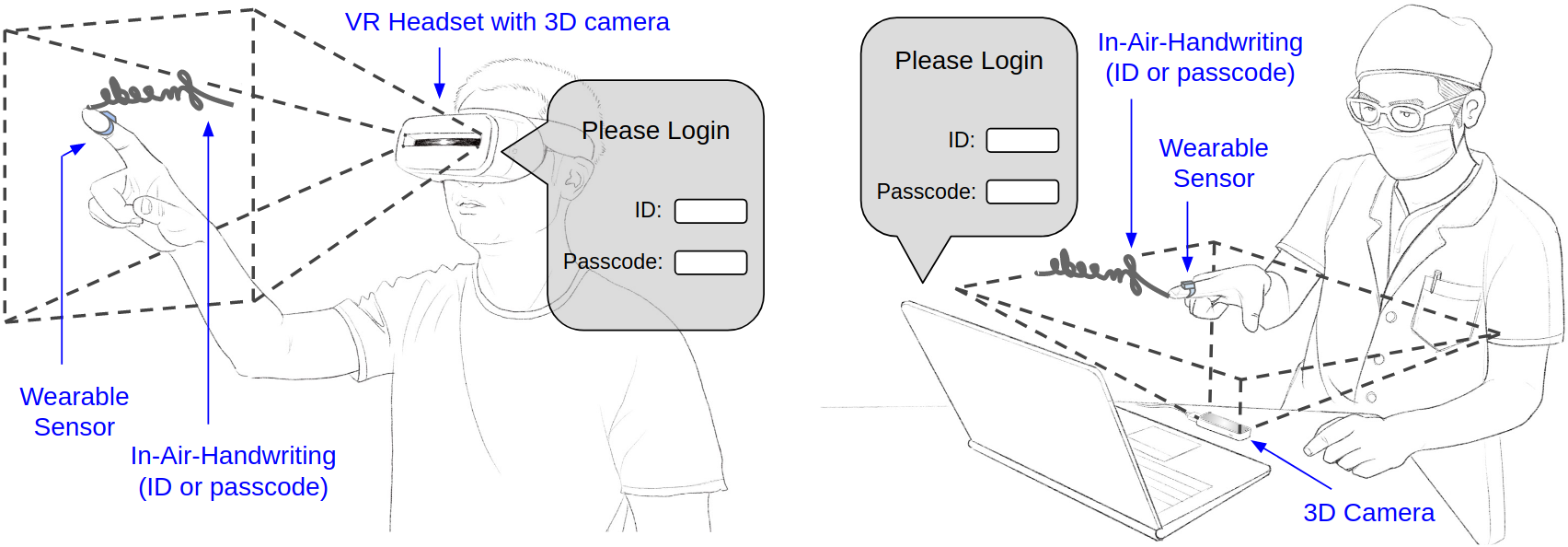}
\caption{User login through gesture interface with hand motion capturing devices of wearable inertial sensor or 3D depth camera under two different scenarios: (left) VR applications with user mobility, (right) operating theater with touchless interface for doctors to maintain high clearliness once the hands are sterilized.}
\label{fig:devices}
\end{figure*}

To address these problems, we propose \textbf{FMCode} (\textit{i.e.,} Finger Motion Passcode), a unified user login framework by writing an ID and a passcode in the air with the index finger, shown in Fig. \ref{fig:devices}. The finger motion is captured by either a wearable inertial sensor or a 3D depth camera and sent to the server as a login request. The user can write very fast, write in a language unable to be directly typed with a keyboard, or just doodle, as long as the finger motion can be easily reproduced in a stable way. Based on our datasets, usually the content of the ID or the passcode is a meaningful string or a shape that is easy to remember, while the writing can be just scribbles unreadable for anyone except the creator. This enables a much larger space for the ID and the passcode than the traditional password consisting of only typed characters. Additionally, because of the difference in writing convention of different people, memorable passcodes are not susceptible to attack as traditional password. For example, the user can draw a five-point star as an ID or a passcode which cannot be typed over the keyboard. Meanwhile, it has been proved that graphical memory is easier to remember and stays longer than passwords typed over the keyboard \cite{GPass}. Due to this unique feature, we call the in-air-handwriting as a ``passcode'' instead of a ``password''. Different from handwriting recognition, our framework does not try to understand each character, but identifies and authenticates users based on multiple factors including both the difference of passcode content and the fineness of handwriting convention. Hence, FMCode has both advantages of password such as revocability and advantages of biometrics such as non-repudiation.

The contributions of this paper are listed as follows.

\begin{enumerate}

\item We present a unified login framework for gesture interface using the finger motion in-the-air, which can perform both user authentication and identification efficiently. Our framework supports gesture interface using either a wearable sensor or a contactless 3D camera.

\item We design a type of feature named temporal local difference and train an ensemblement of Support Vector Machine (SVM) classifiers for each account to improve the discriminant capability. Our method can tolerate minor writing behavior variation and perform reasonably well in the long term.

\item We design a deep Convolutional Neural Network (CNN) to index finger motion signals, which enables user identification with a constant time cost. Moreover, we invent special data augmentation methods to train this network with limited amount of data acquired at registration.
 
\item We achieve state-of-the-art performance on datasets collected by us with two types of gesture input device - a custom data glove and a Leap Motion controller. The dataset contains more than 100 subjects from general public and over 20,000 data points, which is larger than most related work.

\item We study the performance under active spoofing attacks and the long term stability with a duration of four weeks. These scenarios are usually neglected by existing works. Our results show that the FMCode framework has the capability to defend client side spoofing attack and perform persistently in the long term. This reveals great potential of this finger motion based user login method for practical usage.

\end{enumerate}

The structure of the paper is organized as follows. In Section II, we describe the overall system architecture and in Section III we model the signal of the in-air-handwriting motion. Section IV and V explains details of the proposed authentication and identification methods. In section VI we report empirical evaluation results with our datasets, and in section VII, we present user evaluation results. Further discussions are given in section VIII, and the related work is provided in Section IX. Finally, we draw conclusions in section X. More detailed comparison with existing work and alternative technologies are shown in the Appendices.

\begin{figure*}[t]
\centering
\includegraphics[width=7.7in]{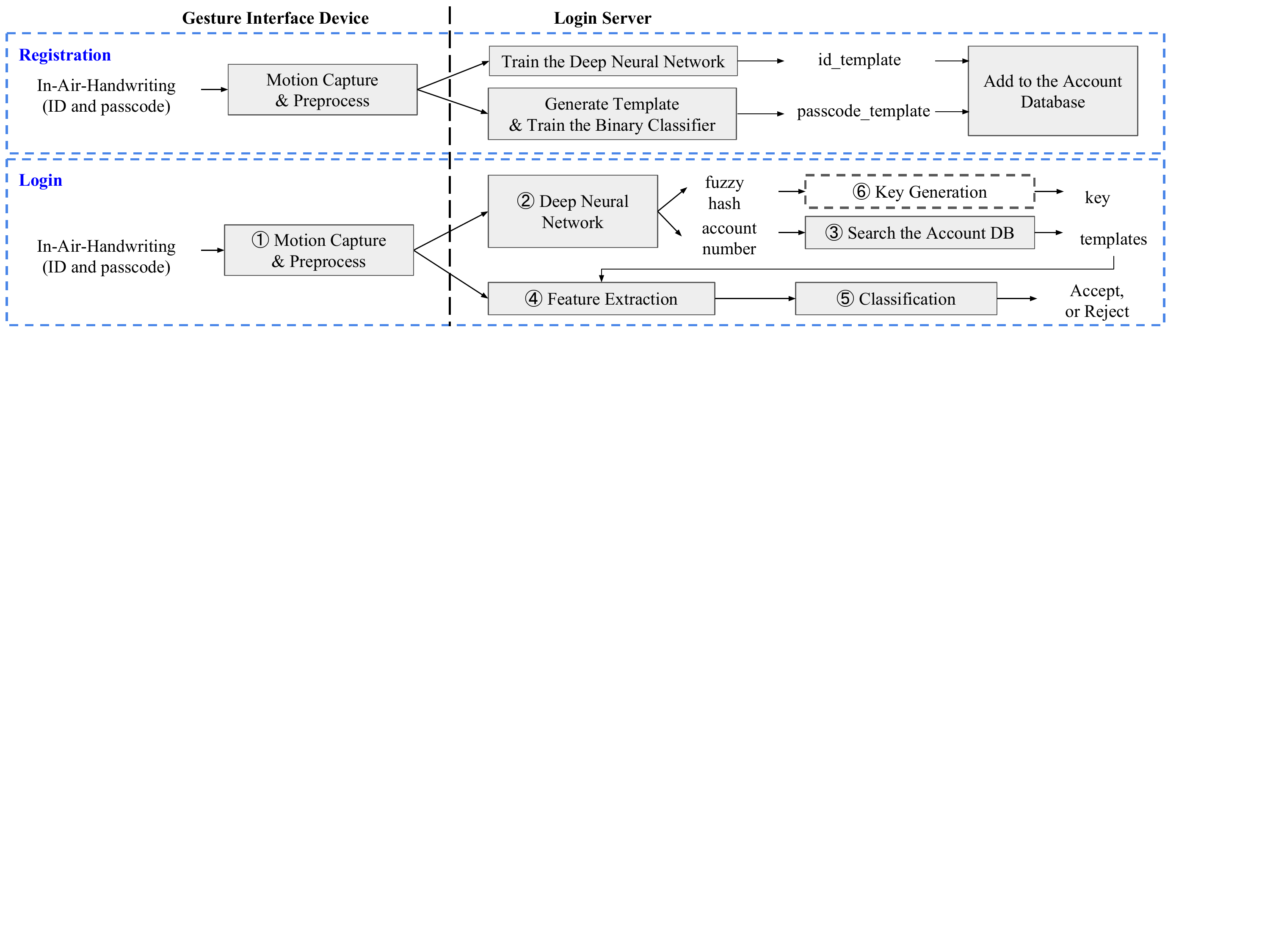}
\vspace{-4.0in}
\caption{System Architecture and Procedures.}
\label{fig:arch}
\end{figure*}

\section{System Model}

\subsection{FMCode Architecture}

Similar to password based login systems, FMCode requires an initial registration comparable to the password ``sign up'' step. At the registration step, the user is required to create an account with a unique ID and a passcode, and write the ID as well as the passcode a few times in the air. After that, the user can sign in to the created account by writing the same ID and passcode in the air.

There are two functional modules in the framework: (a) a gesture user interface device that is equipped with motion capture sensors, and (b) a login server that stores the account database, as shown in Fig. \ref{fig:arch}. On the user side, the finger motion of a piece of the in-air-handwriting of the ID and the passcode is obtained, and two corresponding signals containing the physical states of the hand are generated. Then, the gesture UI device preprocesses the signals and sends them to the login server. On the server side, each account in the database contains a (\textit{account number, id\_template, passcode\_template}) tuple. The account number is a number or a string of characters usually allocated by the server to uniquely index each account. The two templates are generated by the in-air-handwriting signals of the ID and the passcode at registration in order to match with signals in the login request. Once the registration is complete, the login server has the following two main functions that can be used independently or together.

\textbf{1) User Authentication}: given the account number, verify the user identity, and ``accept'' or ``reject'' the login request using a piece of in-air-handwriting of the passcode in the request. The account number may be typed, remembered, recognized by face or other means, or obtained using the identification function detailed below. In this function, the server executes the following three steps: (a) retrieves the \textit{passcode\_template} of the user's account according to the account number from the account database (step 3 in Fig. \ref{fig:arch}), (b) compares the template with the signal in the login request to extract features (step 4 in Fig. \ref{fig:arch}), (c) determines whether this login request is accepted or rejected using a binary classifier trained at registration (step 5 in Fig. \ref{fig:arch}).  

\textbf{2) User Identification}: figure out the account number based on a piece of in-air-handwriting of the ID. As mentioned in the previous function, an account number is required, which is a number or a character string inconvenient to enter through the gesture user interface. Thus, we can ask the user to write the ID in the air. In this function, the server first obtains one or multiple candidate account numbers using a deep CNN (step 2 in Fig. \ref{fig:arch}). Then for each of them, the server runs the same three steps as the authentication to verify each of them by comparing the \textit{id\_template} and the signal of the in-air-handwriting of the ID. Finally, the best matched account number is returned. If all candidate IDs fail the verification, ``unidentified'' is returned. 

Note that identifying the account number does not necessarily mean authenticating the user at the same time because an ID is usually not a secret unlike a passcode. The object of authentication is low error and high security level, while the objective of identification is fast speed with an acceptable accuracy. The login procedure is essentially performing both identification and authentication.  Moreover, the user can explicitly update or revoke his or her FMCode just like updating or resetting a password at anytime. In addition, the server can generate fuzzy hash from the in-air-handwriting of the passcode using a similar deep CNN \cite{FMHash}. This fuzzy hash can be used to further generate cryptographical keys to enable more sophisticated authentication protocols or encrypt the template used in the login procedure in order to minimize the risk of server storage leakage (discussed in section VIII).

\subsection{System Requirements and Application Scenarios}

FMCode is compatible with existing IT infrastructures using password based authentication. On the server side, only software changes are required, including the construction of templates, the implementation of feature extraction algorithms, the classifier and the deep CNN. The requirement of the network between the client and the server is the same as most password-based authentication systems, through the web. On the client side, a motion capture device such as a wearable device or a 3D depth camera is required. However, it should be noted that our login framework leverages the built-in gesture interface of the client machine rather than requires a dedicated device for the login purpose. As long as the finger motion can be captured for ordinary gesture based interaction with the computer, our framework can be deployed. Besides, login through our framework requires the active involvement of a conscious and willing user, rather than presenting a piece of static information such as password or fingerprint. This makes FMCode potentially more secure.  

Our target application scenarios include VR headsets and wearable computers that already have a gesture interface but lack keyboard or touchscreen, as well as scenarios that provide gesture interface but inconvenient for typing such as operating theater or clean room. Our framework can be used for two different purposes. The first one is online user authentication or identification, where the server is remotely connected to the client via the Internet. For example, the user can sign into an online personal account through the gesture interface on a VR headset. The second one is local user authentication or identification, where the client and server reside on the same machine. For example, the user can unlock his or her wearable devices through the gesture interface. In addition, our framework can also be used as a supplementary authentication factor in a Multi-Factor Authentication (MFA) system together with traditional password or biometrics.

\subsection{Attack Model}

FMCode has the same security assumptions as existing gesture-based authentication and identification systems as follows: (1) the device on the user side is secure(\textit{i.e.,} no sniffing backdoor); (2) the authentication server is secure (\textit{i.e.,} it will not leak the stored template to malicious attackers); and (3) the communication channel between the user and the server is secure (\textit{i.e,} no man-in-the-middle attack). These security assumptions are also similar to traditional biometrics and password-based system. Attacks with relaxed security assumptions on (2) and (3) are further discussed in section VIII. Based on the assumptions, we are mainly interested in attacks on the user side, as listed below:
    
\textbf{1) Random guess}, \textit{i.e.,} the attacker tries to enter a user's account by guessing a passcode and signs it on the same gesture interface, without any knowledge of the content of the passcode;
 
\textbf{2) Spoofing attack}, \textit{i.e.,} the attacker knows the content and broad strokes of the passcode of an account and tries to write it in the air through the same gesture interface. This is similar to the case that the attacker sign into the victim's account with the password leaked. 

For the spoofing attack, we assume that the attack source is a human attacker, and the attacker's goal is to sign into the victim's account or be identified as the victim. If the attack is successful, the account owner may suffer from loss of the account or leakage of private information. Though it is generally considered that ID is not a secret, in extreme cases, if the attacker is wrongly identified as the victim, he or she may launch further attacks \textit{e.g.,} phishing other sensitive personal information of the victim. 

\section{In-Air-Handwriting Characterization}

\subsection{Finger Motion Signal Model}

\begin{figure}[!t]
\centering
\includegraphics[width=3.5in]{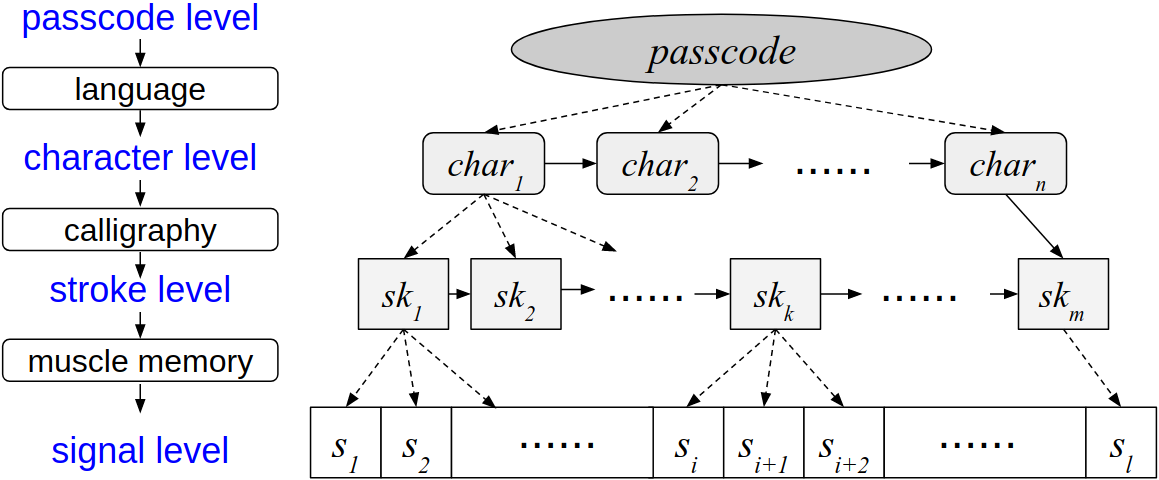}
\caption{The In-Air-Handwriting Signal Model.}
\label{fig:model}
\end{figure}

\begin{figure}[!t]
\centering
\includegraphics[width=3.5in]{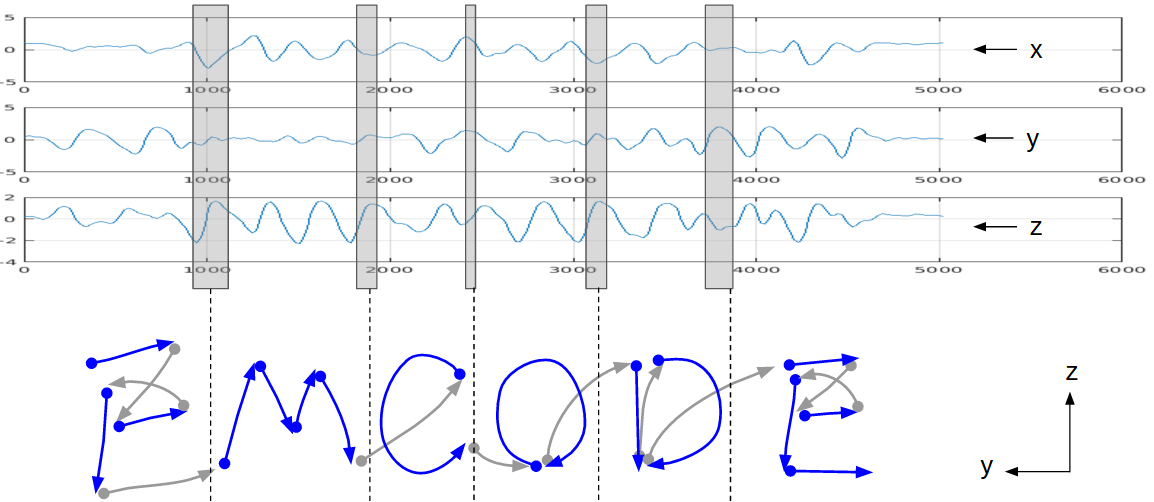}
\caption{Example of 3-dimensional finger motion signal (upper) when writing ``FMCODE'' and the correspondence of signal segments to letters (lower).}
\label{fig:signal}
\end{figure}

Handwriting is closely related to the cognitive process of humans \cite{Itaguchi} in different levels, and hence, we model the in-air-handwriting as a stochastic process in four levels: passcode level, character level, stroke level, and signal level (shown in Fig. \ref{fig:model}). Usually each passcode is a string of meaningful symbols or characters in some language. The passcode is made of strokes defined by calligraphy, and the strokes further determine the hand motion of the writing through the muscle memory. The hand motion is captured as a series of samples of physical states of the hand and fingers. Here the passcode, characters, and strokes can be regarded as hidden states in a stochastic process, and only the signal samples are the observations. In general the in-air-handwriting process does not satisfy the Markov property (\textit{i.e.,} signal samples are correlated), and the mapping between signal samples and strokes are not fixed due to the minor variations of the writing speed and amplitude. However, the inherent process in the brain of generating hand motion by writing is acquired and reinforced when a person learns how to write \cite{Klemmer}, which indicates that the writing behavior is bound to individuals and persistent in the long term, as handwriting signature has been widely used for identity verification for a long time.

\begin{figure}
\centering
\includegraphics[width=3.5in]{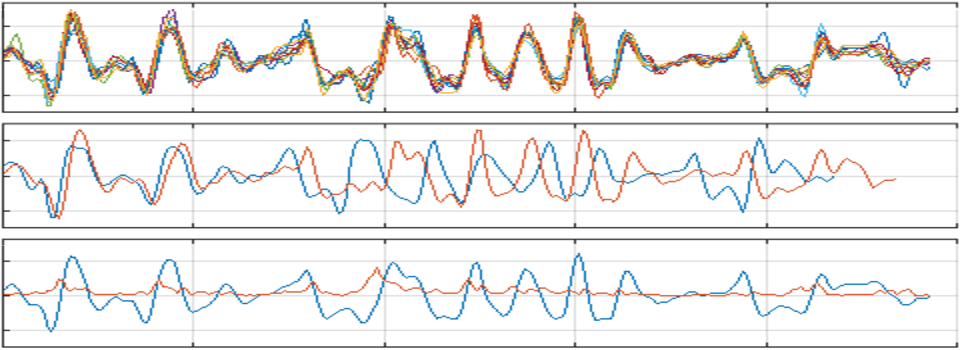}
\caption{Example of 10 aligned finger motion signals (upper), 2 unaligned signals (middle), and the generated template (lower, where $T$ is shown as the blue line, $C$ is shown as the red line, and $C$ is enlarged by two to show significance). Only the first sensor axis is shown. Best viewed with color.}
\label{fig:temp_example}
\end{figure}

We use a vector series $\mathbf{s} = (\mathbf{s}_1, \mathbf{s}_2, ..., \mathbf{s}_l)$ to denote the finger motion signal with $l$ samples, where $\mathbf{s}_i = (s_{i1}, s_{i2}, ..., s_{id})$ represents an individual sample obtained by the sensor with $d$ axes. For example, if the signal is obtained from a wearable device with an inertial sensor, each sample $\mathbf{s}_i$ may have three axes representing the acceleration of the fingertip of the right hand along the x, y and z-axes, and the whole signal $\mathbf{s}$ may have 250 samples at 50 Hz. Assume the signal $\mathbf{s}$ is aligned in a fashion that the writing speed variation is removed, it can be decomposed as 
$$\mathbf{s}_i = \mathbf{t}_i + \mathbf{e}_i, \quad\quad \mathbf{e}_i \sim N(0, \mathbf{\Sigma}_i),$$
\noindent where $\mathbf{t}_i$ is a constant vector determined by the content of the in-air-handwriting, and $\mathbf{e}_i$ is a vector of Gaussian random variables caused by the sensor noise and unintentional small hand movements. Since $\mathbf{e}_i$ is from different orthogonal sensor axes, we assume these axes are independent, \textit{i.e.,} $\mathbf{\Sigma}_i = \boldsymbol{\sigma}_i I$, where $\boldsymbol{\sigma}_i = (\sigma_{i1}, \sigma_{i2}, ..., \sigma_{id})$. An approximation of $\mathbf{t}_i$ and $\boldsymbol{\sigma}_i$ can be obtained by the signals $\{\mathbf{s}^1, \mathbf{s}^2, ..., \mathbf{s}^K\}$ at registration, 
$$\hat{\mathbf{t}}_i = \frac{1}{K} \sum_{k=1}^K\mathbf{s}_i^k, \quad\quad \hat{\boldsymbol{\sigma}}_i = \frac{1}{K-1}\sum_{k=1}^K(\mathbf{s}_i^k - \hat{\mathbf{t}}_i).$$
Here $\hat{\mathbf{t}}_i$ is the Maximum Likelihood Estimation (MLE) of $\mathbf{t}_i$ and $\hat{\boldsymbol{\sigma}}_i$ is the unbiased MLE of $\boldsymbol{\sigma}_i$. For each account, $\hat{\mathbf{t}}$ is stored as the $id\_template$ or the $passcode\_template$, depending on whether $\mathbf{s}^k$ is obtained by writing the ID or the passcode. $\hat{\boldsymbol{\sigma}}$ is also stored together with the template to indicate the template uncertainty. The aligned signal set $\{\mathbf{s}^k\}$ can be obtained by aligning each raw signal at registration to the first signal. An example is shown in Fig. \ref{fig:temp_example}. In our framework, alignment is made by the Dynamic Time Warping (DTW) algorithm.

\subsection{In-Air-Handwriting Signal Preprocessing}

As mentioned previously, there are also user behavior uncertainty in the writing posture and magnitude. To minimize the influence of such uncertainty, the following preprocessing steps are applied on the client side.

\textbf{Step 1) State estimation}: derive the indirect dynamic states from the raw sensor output and fix any missing signal samples due to the limited capability of the motion capture device. For the wearable device with inertial sensors, we derive the absolute orientation of the index finger relative to the starting position \cite{AHRS}. For the 3D camera device, we derive the velocity and acceleration for each sample from the position difference. The position trajectory and posture are estimated from the depth image frames by the 3D camera device itself.

\textbf{Step 2) Trimming}: throw away the sample at the start and the end of the signal where the hand does not move intensively. In practice, the signal can be trimmed in a more progressive way because we observed that the user behavior has larger uncertainty at the beginning and the end.

\textbf{Step 3) Low-Pass filtering and resample}: remove the high-frequency components above $10$ Hz because it is commonly believed that a person cannot generate finger movements faster than $10$ Hz. Then the signal is resampled at 50 Hz to remove influence on the variation of sampling rate.

\textbf{Step 4) Posture normalization}: translate the coordinate system to make the average pointing direction of the hand as the x-axis in order to remove the influence of posture variation respect to the motion capture device. For the data glove with inertial sensors, we also remove the influence of the gravity on the acceleration axes.

\textbf{Step 5) Amplitude Normalization}: normalize the data of each sensor axis individually, \textit{i.e.,} $s_{ij} \leftarrow (s_{ij} - \mu_{j}) / \sigma_{j}$
where 
$\mu_{j} = mean(s_{1j}, ..., s_{lj})$,    $\sigma_{j} = std(s_{1j}, ..., s_{lj}).$

\section{User Authentication}

The task of authentication is essentially a binary classification problem. Our design goal is to build a data-driven model that can optimally distinguish the signals from legitimate users and attackers. Given an account A and an authentication request with signal $\mathbf{s}$ obtained from the in-air-handwriting of the passcode (referred as the passcode signal), we define the following classes. 

\begin{enumerate}
\item If $\mathbf{s}$ is generated by the owner of account A writing the correct passcode, $\mathbf{s}$ is from the ``\textbf{true-user}'' class; 
\item if $\mathbf{s}$ is generated by any user writing an incorrect passcode, $\mathbf{s}$ is from the ``\textbf{guessing}'' class (which means the writer does not know the passcode content); 
\item if $\mathbf{s}$ is generated by an imposter writing the correct passcode of account A, we define that $\mathbf{s}$ is from the ``\textbf{spoofing}'' class (which means the attacker knows the passcode content).
\end{enumerate}

The ``guessing'' and ``spoofing'' classes are collectively called the ``\textbf{not-true-user}'' class. Our authentication method is based on the signal model explained previously -- signals generated by the same user writing the same passcode have similar shape if they are aligned because they contain the same sequence of strokes. Hence, we define a temporal local distance feature that measures the difference of the signals locally in stroke level. Moreover, we also design a method to generate multiple feature vectors from just one pair of signal and template to overcome the shortage of training data at registration. Furthermore, we use an ensemblement of SVM classifiers for each account to distinguish signals from the ``true-user'' class and ``not-true-user'' class to maintain a stable long term performance.

\subsection{Feature Extraction}

Given an account A, consider $\mathbf{s}$ is the passcode signal in an authentication request, and $\hat{\mathbf{t}}$ is the $passcode\_template$ of account A constructed at registration with uncertainty $\hat{\boldsymbol{\sigma}}$. The temporal local distance features are extracted as follows. 

\textbf{Step 1)} Align $\mathbf{s}$ to $\hat{\mathbf{t}}$ using DTW, so that the aligned $\mathbf{s}$ will have the same length $l$ as $\hat{\mathbf{t}}$. 

\textbf{Step 2)} Calculate the distance $\mathbf{d}_i = abs(\mathbf{s}_i - \hat{\mathbf{t}}_i)$, where $abs()$ is the element-wise absolute function.

\textbf{Step 3)} Segment $\mathbf{d}$ into $H$ local windows, and each window has length $W = l / H$, \textit{i.e.,} regroup $\mathbf{d}$ as $(\mathbf{D}_1, \mathbf{D}_2, ..., \mathbf{D}_H)$, where $\mathbf{D}_j = (\mathbf{d}_{j \times W + 1}, \mathbf{d}_{j \times W + 2}, ..., \mathbf{d}_{j \times W + W})$. 

\textbf{Step 4)} Randomly pick $T$ different local windows as a window set $\{j_1, j_2, ..., j_T\}$, then randomly select a local distance feature from each window to form a feature vector $\mathbf{x} = (\mathbf{x}_{j1}, \mathbf{x}_{j2}, ..., \mathbf{x}_{jT})$, where each element $\mathbf{x}_j$ is chosen from $\mathbf{D}_j$. Here $\mathbf{x}$ is defined as the temporal local distance feature. For example, assume $\mathbf{d}$ has 10 samples, segmented to five windows, and we can randomly pick two windows (i.e., $l = 10, H = 5, W = 2, T = 2$). Consider we pick the third window ($\mathbf{d}_{5}$ to $\mathbf{d}_{6}$) and the fifth window ($\mathbf{d}_{9}$ to $\mathbf{d}_{10}$), then we can form a feature vector by randomly choosing one sample from each window, such as $(\mathbf{d}_{6}, \mathbf{d}_{9})$.

\textbf{Step 5)} Given a certain window set, step 4 can be repeated multiple times to generate multiple feature vector from one pair of signal and template. Especially, we can regard $\mathbf{d}_i$ as a Gaussian random variable and draw samples from the distribution $\mathbf{d}_i \sim N(abs(\mathbf{s}_i - \hat{\mathbf{t}}_i), \hat{\boldsymbol{\sigma}}_i)$ in step 4. This technique allows us to augment the training data from the limited signals with ``true-user'' label at registration.

\begin{figure*}[]
\centering
\includegraphics[width=7in]{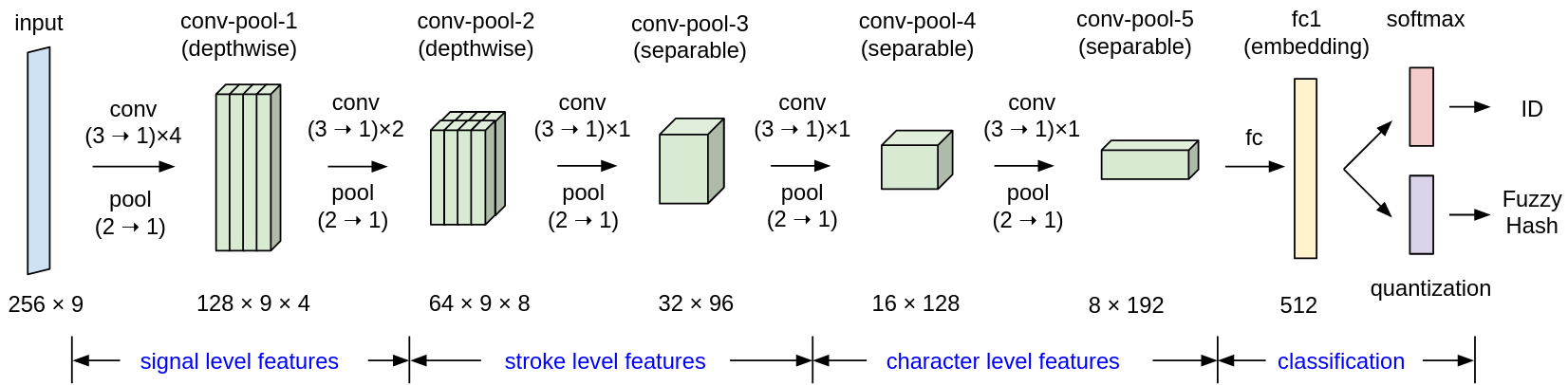}
\caption{Convolutional neural network for user identification.}
\label{fig:nn}
\end{figure*}

\subsection{Binary Classification for Authentication}

The SVM classifier is a binary classifier with linear decision boundary in the feature space that can be trained efficiently even with limited amount of data and high dimension of features. These characteristics are suitable for the authentication task. Given a training dataset with data points $\{(\mathbf{x}^1, y^1), (\mathbf{x}^2, y^2), ..., (\mathbf{x}^n, y^n)\}$, where $\{\mathbf{x}^i\}$ are the feature vectors and $\{y^i\}$ are binary class labels from $\{-1, +1\}$, SVM seeks a hyperplane $f(\mathbf{x}) = \mathbf{w} \mathbf{x} + b = 0$ to maximize the separation of the data points of the two classes. Training SVM is equivalent to solving a quadratic programming problem, which can be done efficiently. However, since the decision boundary is very simple, naively applying SVM on $\mathbf{d}$ obtained from limited signals at registration would still suffer from the curse of dimensionality problem and lead to poor long term stability. Hence, we train an ensemble of SVM as follows.

Consider we are registering the account A and the template $\hat{\mathbf{t}}$ is constructed from signals $\{\mathbf{s}^1, \mathbf{s}^2, ..., \mathbf{s}^K\}$. At registration, the server builds $M$ SVM classifiers for account A, with a distinct set of $T$ windows $\{j_1, j_2, ..., j_T\}$ randomly picked for each classifier. To train a single classifier, first the server extracts feature vectors from those $K$ registration signals of account A, and assigns them the label $y = +1$ (\textit{i.e.,} ``true-user'' class). Then the server extracts feature vectors from those registration signals of other accounts except the account A, and assigns them the label $y = -1$ (\textit{i.e.,} ``guessing'' class). Usually there are more training data of the ``guessing'' class than necessary. Thus, only a portion is needed (usually around one thousand). After the feature vectors and lables of both classes are ready, an SVM is trained using Sequential Minimal Optimization (SMO). Finally, once all $M$ classifiers are trained, the server stores the model parameters $\mathbf{w}$, $b$, and the set of windows $\{j_1, j_2, ..., j_T\}$ of each classifier in the account database together with the template $\hat{\mathbf{t}}$.

When signing into account A, given a signal $\mathbf{s}$ in the authentication request, the server extracts feature vectors for each SVM classifier using the stored information, and predicts a score $f(\mathbf{x}) = \mathbf{w} \mathbf{x} + b$. Since multiple feature vectors can be extracted with one set of windows, the server can obtain multiple scores from a single $\mathbf{s}$ and average them. Once the scores of all classifiers are ready, they are further averaged to produce a single distance score, \textit{i.e.,} $score(\mathbf{s}) = average(\{f(\mathbf{x)}\})$. Finally, this score is compared with a pre-set decision threshold. If $score(\mathbf{s}) < threshold$, the authentication request with signal $\mathbf{s}$ is accepted, otherwise, it is rejected.

The aim of the feature extraction method and classifier ensemblement is to achieve a better separation of signals from different classes in the feature space and maintain a stable performance in the long term. If a signal is from the ``true-user'' class, it should have a small score because similar shape between the signal and the template leads to smaller $\mathbf{d}$. While signals from the ``not-true-user'' classes should have a larger score caused by large values of elements of $\mathbf{d}$ due to shape differences, whose origin is the different content or different writing convention. However, the distance in sample level has uncertainties because of the minor variation of writing behavior for the same user writing the same content. Misclassification caused by such uncertainties may happen quite often if we blindly sum the sample level distance, as the plain DTW algorithm. Instead, we group local samples into segments which roughly map to the strokes, and hence, our method can tolerate the signal level uncertainties by comparing subset of strokes instead of samples. The final score of the ensemblement of classifiers is essentially a weighted sum of the sample-wise distance $\mathbf{d}$, where the trained weights help select those segments with less uncertainty and more consistency. In our framework, $H$, $T$ and $M$ are a system-wide parameter. $H$ is usually chosen from 20 to 80 to approximate the number of strokes of a in-air-handwriting passcode. $T$ is usually empirically chosen based on the passcode length to avoid the curse of dimensionality. $M$ is determined as a trade-off between the computation time and authentication accuracy. In an extreme case, we can make $T = H$ and $M = 1$, which means only a single SVM is used to draw a decision boundary in a very high dimensional feature space (potentially this dimension can reach several hundred or even exceed one thousand). This may cause classifier stability issues in the long term because some local features may be wrongly considered to be consistent due to the limited amount of training data.

\section{User Identification}

Different from authentication, the task of identification is essentially a multi-class classification problem, which must be done efficiently without query the account database linearly. Deep CNN is a type of neural network consisting of cascaded convolutional layers and pooling layers. The most attractive capability of deep CNN is that it can learn and detect features from low-level to high-level automatically by optimizing a loss function iteratively. This characteristic is crucial to solve very complicated pattern recognition problems, most notably the image classification \cite{krizhevsky2012imagenet, he2016deep, chollet2016xception}. As illustrated previously in the signal model, a piece of in-air-handwriting contains hierarchical features in different abstract level, and hence, a deep CNN is suitable to our objective of mapping a signal $\mathbf{s}$ of the in-air-handwriting of the ID (referred as the ID signal) to its corresponding account number. However, deep CNN has not been used in in-air-handwriting signal based user identification since the features expected to be learned in 3D handwriting signal are fundamentally different from 2D image, which requires the following special treatment.

The deep CNN in our framework contains five convolutional-pooling layers, one fully-connected layer, and one softmax layer, as shown in Fig. \ref{fig:nn}. The raw signal is first preprocessed and stretched to a fixed length of 256 elements through linear interpolation, in order to be fed into the CNN. For example, if the number of sensor axes is 9, the input is a 256 by 9 matrix, where each sensor axis is regarded as an individual channel. For each convolutional-pooling layer, we apply a convolutional kernel of size 3 on all the channels of the previous layer, and a 2-by-1 maxpooling on the output of the convolutional layer for each channel. The first two convolutional layers utilize depthwise convolution \cite{chollet2016xception} which detects local features individually on each channel since these channels contain different physical states in orthogonal axes. The later three convolutional layers utilize separable convolution \cite{chollet2016xception} which associates low level features on all channels to construct high level features. For example, each neuron in the third conv-pool layer has a receptive field of 16 samples in the original signal, which is roughly corresponds to one stroke. There are 96 filters in this layer which can map to 96 different types of features in the stroke level. These features can capture different types of basic finger movement when writing a single stroke, including straight motion in different directions, and sharp turning between adjacent strokes. Hence, the output of this layer is a 32 by 96 matrix indicating the presence and intensity of a certain type of stroke (among all 96 types) at a certain place of the signal (among 32 slightly overlapping segments). Similarly, the fourth and fifth conv-pool layers are designed to detect the presence of certain characters and phrases. Finally, a fully connected layer runs classification on the flattened high level features and generates the embedding vectors, and the softmax layer maps the embedding vectors to probability distribution of class labels (\textit{i.e.,} the account numbers).

A major challenge of training the CNN is the limited amount of data at registration, which leads to overfitting easily. To overcome this hurdle, we augment our training dataset with the following three steps. First, given $K$ signals $\{\mathbf{s}^1, \mathbf{s}^2, ..., \mathbf{s}^K\}$ obtained at registration, for each $\mathbf{s}^k$ in this set, the server align all the other signals to $\mathbf{s}^k$ to create $K - 1$ new signals. Second the server randomly picks two aligned signals and exchanges a random segment. This can be done many times to further create many new signals. Third, for each newly created signal, we randomly perturb a segment both in time and amplitude. Besides of the data augmentation, we apply dropout \cite{srivastava2014dropout} on the fully-connected layer.

To predict the account number of an input signal, the server simply chooses the most probable class or top-k most probable classes in the output probability distribution. However, blindly believing the predicted account number of the CNN may render the server extremely vulnerable to spoofing attacks because the decision is based on the presence of certain strokes detected by the feature layers, and a spoofed signal generating by writing the same content as the genuine signal naturally has the majority of the strokes. As a result, in our framework the server performs an additional step to verify the identity, using the same procedure as the authentication. Instead of passcode signal, here the server compares the ID signal and the \textit{id\_template} of the account corresponding to each candidate account number. Finally, the server returns the best matched account number or ``unidentified'' if the scores of all account are above the threshold. 

\section{Experimental Evaluation}

\subsection{Data Acquisition}

\begin{figure}[!t]
\centering
\includegraphics[width=3in]{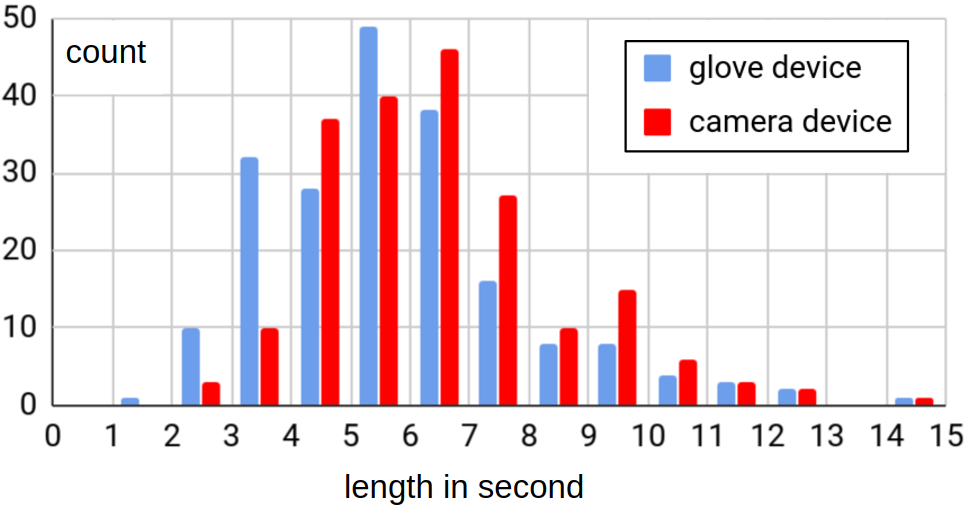}
\caption{Distribution of passcode lengths of the signals in dataset 1.}
\label{fig:len_dist}
\end{figure}

To evaluate our framework, we build a prototype system with two types of gesture input device. The first device is a custom made data glove with an inertial measurement unit (IMU) on the tip of the index finger (referred as the glove device). Such an IMU has been widely used in the hand-held remote controller of current VR game console and smart TV, and it is relatively inexpensive ($<$\$10). The glove also has a battery-powered microcontroller on the wrist, which collects the IMU data at $50 Hz$, and runs the state estimation preprocessing step. The output signal of this device contains a series of three Euler angles in $\pm 180\degree$, tri-axis acceleration in $\pm 4g$, and tri-axis angular speed in $\pm 2000dps$. The second device is the Leap Motion controller (referred as the camera device), which is an inexpensive ($\sim$\$80) commercial-off-the-shelf 3D camera. It contains a stereo infrared camera that can capture depth image at $55 Hz \sim 110 Hz$ with an average $135\degree$ Field of View (FOV) and a range of $60 cm$. The Leap Motion controller has its own processor that can estimate the hand skeleton of each frame and obtain the 3D coordinate of each joint of the hand. The client machine runs the state estimation and other preprocessing steps, and outputs signal of a series of 3D position, velocity, and acceleration. 

Although our prototype system uses these two specific devices for proof of concept and evaluation, the proposed framework does not necessarily depend on these two devices. It should work with any similar device that can return samples of physical states of the fingers and the hand with a reasonable range, resolution and sampling rate. For example, there are other gloves \cite{Glove} and rings \cite{Ring} with inertial based motion tracking, and there are open source software \cite{OpenPose} available that can estimate hand skeleton from 2D image or 3D depth image. We assume these devices are part of the gesture user interface for ordinary human-computer interaction, not specialized device dedicated for the login purpose. 

We built our own datasets as follows:

\textbf{1)} We invited 105 and 109 users to participate in the data collection with the glove device and the camera device respectively. 56 users used both devices. Each user created two distinct strings. For each string, the user wrote it in the air for 5 times as registration, and after a rest, they wrote it 5 times again as login. The distribution of lengths of all strings are shown in Fig. \ref{fig:len_dist} (users write a little slower with the camera device due to its limited FOV and range).

\textbf{2)} For each of the string in the first dataset, 7 impostors spoofed it for 5 times (due to the size of the first dataset there are more than 7 impostors in total). Here ``spoof'' means that the impostors know the content of the string, and try to write it using the same motion capture device as that used by the spoofing target user.

\textbf{3)} We asked 25 users participating in the first dataset to write the two created strings in multiple sessions. For each of the string, besides the 5 times at registration, the user wrote it 5 times as one session, two or three sessions a week on different days, for a period of 4 weeks (10 sessions in total).

\textbf{4)} Among the users who participate in the data collection of both devices in the first dataset, 28 of them filed a questionnaire on the usability of our prototype system (detailed in section VII)

The datasets and data acquisition method are based on our previous works \cite{FMCode-IJCB, FMCode-ICB}. However, our datasets are larger with two types of devices. The participating users are from the general public including both office workers and non-office workers with an age range of 17 to 65.

%We obtained a permission from our institutional ethics review board (reference number not disclosed here due to anonymous submission policy), and we made agreements with each of the participants that the data would only be used for research purpose in anonymous ways. 

At the beginning of the data collection, we briefly introduced our system to the users and informed them that the in-air-handwriting is for login purpose. The users are allowed to practice to write in the air for a few times. Most of them can understand the idea easily and get prepared within a minute. During the data collection, the user can voluntarily abort the writing and the incomplete data is discarded. Only one data glove and one Leap Motion controller are used and the data collection processes with the two devices are separate. All users write with the right hand and wear the glove on that hand. The Leap Motion controller is placed on a table or a side table. Both devices are connected to a laptop as the client machine. For the first dataset, there is no constraint on the content of the string created by the user except distinctiveness. Also there is no constraint on the writing convention. For example, the user can write in various direction, stack every character on the same place, write while standing or sitting, with elbow supported or not supported. Most users write very fast and their writing is illegible, like the way of signing a signature. Since the strings in the first dataset are distinct, we use them either as a passcode for authentication or an ID for identification.

\subsection{Authentication Experiments and Results}

For authentication, each individual string is considered as the passcode of an account, and in total there are 210 and 218 accounts with the glove and camera device respectively. We run the following procedures with the 64 window ($H = 64$), 16 local features for each individual classifier ($T = 16$), and 32 classifiers as an ensemblement ($M = 32$).

\textbf{1)} We follow the registration process to create all the accounts, construct $passcode\_template$ and train the classifier for each account. For the five signals for registration, all of them are used to construct the template and train the SVM classifier. Thus, For each account there are five training signals with ``true-user'' label, and $5 \times (210 - 1)$ or $5 \times (218 - 1)$ training signals with ``not-true-user'' label (\textit{i.e.,} the training signals of other accounts).

\textbf{2)} We use each $\mathbf{s}$ of the five testing signals of the same account as an authentication request (\textit{i.e.,} the ground truth label of $\mathbf{s}$ is ``true-user''). If a signal from the ``true-user'' class is misclassified to ``not-true-user'' class, the result is a False Reject (FR); otherwise it is a True Accept (TA). 

\textbf{3)} We use each $\mathbf{s}$ of the five testing signals of an account as an authentication request to all other accounts (\textit{i.e.,} the ground truth label of $\mathbf{s}$ is ``guessing''). If a signal from the ``guessing'' class is misclassified to the ``true-user'' class, the result is defined as a False Accept (FA), otherwise it is a True Reject (TR).

\textbf{4)} We use each $\mathbf{s}$ of the five spoofing signals in the dataset 2 as an authentication request to the spoofing target account (\textit{i.e.,} the ground truth label of $\mathbf{s}$ is ``spoofing''). If a signal from the ``spoofing'' class is misclassified to the ``true-user'' class, the result is defined as a Successful Spoof (SS), which is considered as a special case of FA.

\begin{figure}[]
\begin{center}
\includegraphics[width=3.5in]{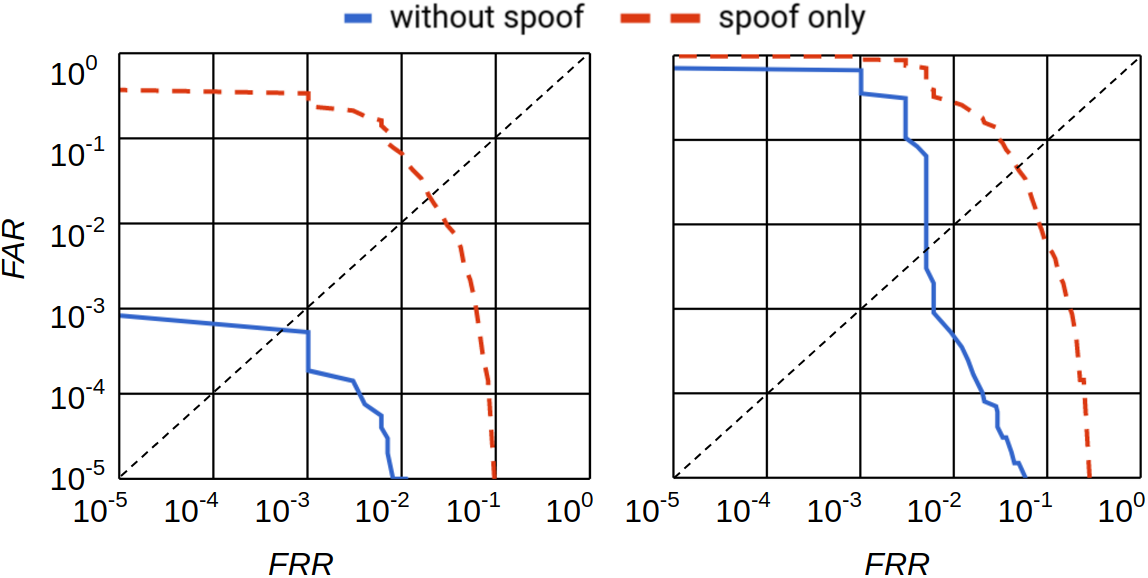}
\end{center}
   \caption{ROC with the glove device (left) and the camera device (right). ``without spoof'' means the plot of FAR against FRR, and the line of ``spoof only'' refers plotting FAR$_{spoof}$ against FRR}
\label{fig:roc}
\end{figure}

The main evaluation metrics are False Reject Rate (FRR) and False Accept Rate (FAR), which are the the portions of false rejects and false accepts in all authentication requests respectively, formally defined as follows:

$$FRR = \frac{1}{kn}\sum\limits_{i=1}^{n} \#\{FR_i\},$$
$$FAR = \frac{1}{kn(n - 1)}\sum\limits_{i=1}^{n}\sum\limits_{j=1, j \ne i}^{n} \#\{FA_{ij}\},$$
$$FAR_{spoof} = \frac{1}{kmn}\sum\limits_{i=1}^{n}\sum\limits_{j=1}^{m} \#\{SS_{ij}\},$$

\noindent where $n$ is the total number of accounts, $k$ is the number of testing authentication requests for each account, $m$ is the number of impostors. $\#\{FR_i\}$ is the number of FR of the $ith$ account, $\#\{FA_{ij}\}$ is the number of FA for $ith$ account using the signals of $jth$ account as the authentication requests, and $\#\{SS_{ij}\}$ is the number of successful spoof for the $ith$ account by the $jth$ impostor. Equal Error Rate (EER) is defined as the rate where FRR is equal to FAR. FAR10K and FAR100K denoets the FRR when FAR is $10^{-4}$ and $10^{-5}$, respectively. ZeroFAR denotes the FRR when FAR is zero. We varying the decision threshold to change the amount of FR and FA, and results are shown in the Receiver Operating Characteristic (ROC) curve in Fig. \ref{fig:roc}. The results are shown in Table \ref{tb:auth_results}, with comparison to plain DTW method on the same dataset with the same signal preprocessing techniques using the same template. These evaluation metrics are widely used in traditional biometric authentication systems such as fingerprint \cite{Fingerprint}. For comparison, we mainly use the EER since it is a single number that captures the general performance, and for practical usage, FAR10K is more important.

Compared with plain DTW algorithm which is used in many related works, our method has a significant performance improvement. We believe DTW is a good alignment algorithm but not necessarily a good matching algorithm. Our method treats different local distance with different importance and considers different segments of signal individually, while DTW uses a sum of element-wise distance that may propagate locally miss matched sample pair to the final result even for signals in the ``true-user'' class. Besides, DTW treats every segment of the signal equally, while some segments can be more distinctive for different strings, \textit{e.g.,} the second half is more important for the signals generated by writing ``PASSCODE'' and ``PASSWORD''. In general the data obtained by the glove device has higher consistency because it does not restrict the user to write in a certain area. Also the signal quality is better with the glove device, while signals with the camera device can contain missing samples or wrongly estimated hand postures. Hence, our prototype with the glove device has better performance.

\begin{table}[]
\centering
\caption{Empirical results of authentication}
\label{tb:auth_results}
\begin{tabular}{|l|l|l|l|l|l|}
\hline
Metric        & EER   & \begin{tabular}[c]{@{}l@{}}EER \\ (spoof)\end{tabular} & \begin{tabular}[c]{@{}l@{}}FAR\\ 10K\end{tabular} & \begin{tabular}[c]{@{}l@{}}FAR\\ 100K\end{tabular} & \begin{tabular}[c]{@{}l@{}}Zero\\ FAR\end{tabular}    \\ \hline
Ours (glove)  & 0.1\% & 2.2\%                                                  & 0.4\%                                             & 0.9\%                                              & 1.0\%                                                \\ \hline
DTW (glove)   & 0.6\% & 3.9\%                                                  & 3.0\%                                             & 9.2\%                                              & 13.1\%                                                \\ \hline
Ours (camera) & 0.5\% & 4.4\%                                                  & 1.9\%                                             & 4.6\%                                              & 5.6\%                                                \\ \hline
DTW (camera)  & 2.1\% & 7.7\%                                                  & 15.9\%                                             & 28.5\%                                              & 30.8\%                                                \\ \hline
\end{tabular}
\end{table}

\subsection{Identification Experiments and Results}

\begin{figure*}[t]
\centering
\includegraphics[width=7in]{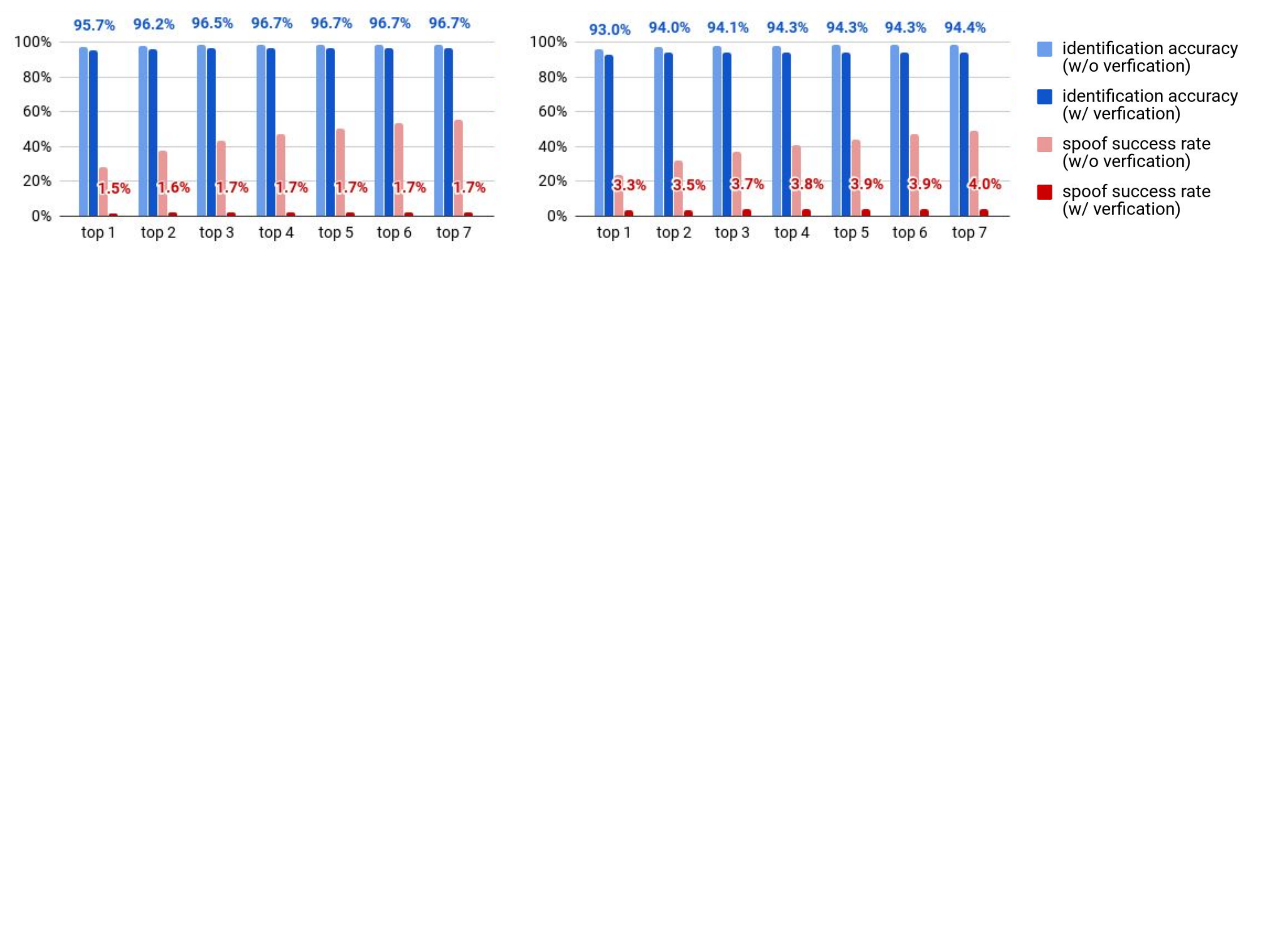}
\vspace{-4.0in}
\caption{Identification performance with the glove device (left) and the camera device (right). The annotation shows results with identity verification.}
\label{fig:ident_results}
\end{figure*}

In the identification experiments, each individual string is considered as the ID associated to an account. For the five signals at registration, we augment them to 125 signals to train the CNN. The activation function we use in the convolutional layer and fully-connected layers is a rectified linear unit (ReLU). The whole network is trained for 20 epochs with Adam algorithm \cite{kingma2014adam} with a learning rate of 0.001. We use a combination of cross-entropy loss and the center loss \cite{wen2016discriminative} as the optimization target. The experiment procedures are as follows.

\textbf{1)} First, we follow the registration process to create all the accounts, train the CNN, construct the template and train the SVM ensemblement.

\textbf{2)} Next, we use each $\mathbf{s}$ of the five testing signals of each account as an identification request and run the identification process. If $\mathbf{s}$ is from account A and the predicted account number is A, it is a correct identification; otherwise it is an incorrect identification. 

\textbf{3)} After that, we use each $\mathbf{s}$ of the five spoofing signals in dataset 2 as an identification request. If $\mathbf{s}$ is a spoofing attack targeting account A and the predicted account number is A, it is a successful spoof; otherwise it is an unsuccessful spoof.

We also run these experiments without the identity verification. In this case, if the $\mathbf{s}$ is a testing signal from account A or a spoofing attack targeting account A, and if the top-k candidate account numbers predicted by the CNN contains A, it is a correct identification or a successful spoof respectively.

It is commonly believed that ID is not a secret and the identity verification should not be too strict to hurt usability. Thus, we choose the threshold of the identity verification with a value that achieves the EER for spoofing only data. We vary $k$ from 1 to 7 and show the results in Fig. \ref{fig:ident_results}. In general, increasing the number of candidates helps identification accuracy, at a marginal cost of slightly increased spoof success rate. However, if identity verification is skipped, spoof success rate will have a significant increase, which renders the system vulnerable or even useless. The main cause is that the CNN learns features for distinguishing difference stroke combinations instead of distinguishing fine differences in the signal segments of the same stroke. Also in practice, there is no spoofing data available at registration time to train the CNN. Essentially, the CNN also serves as an index for all accounts in the database, which locates the probable accounts given a signal instead of search it exhaustively. With exhaustive search, our prototype system can achieve 99.9\% accuracy with the glove device and 99.5\% accuracy with the camera device. However, it takes more than one second to exhaustive search on a database containing about 200 accounts. The time consumption is mainly caused by accessing the stored template as well as aligning the signal, and it will grow linearly with the number of accounts in the database. More details on running time are shown in Appendix A.

\begin{figure*}[t]
\centering
\includegraphics[width=7in]{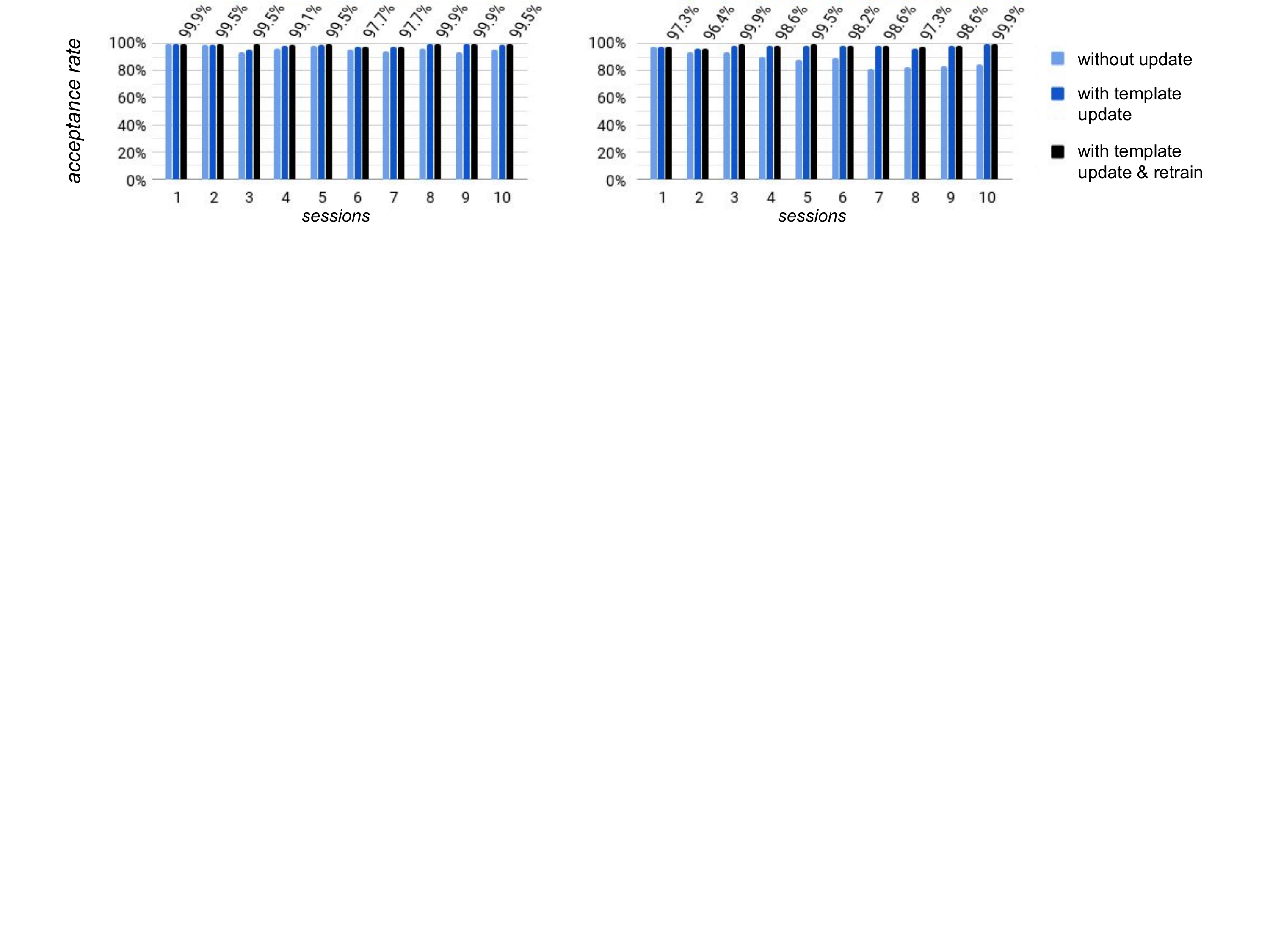}
\vspace{-4.0in}
\caption{Authentication permanence performance with the glove device (left) and the camera device (right). Annotation shows results with update and retrain.}
\label{fig:auth_permanence}
\end{figure*}

\begin{figure*}[t]
\centering
\includegraphics[width=7in]{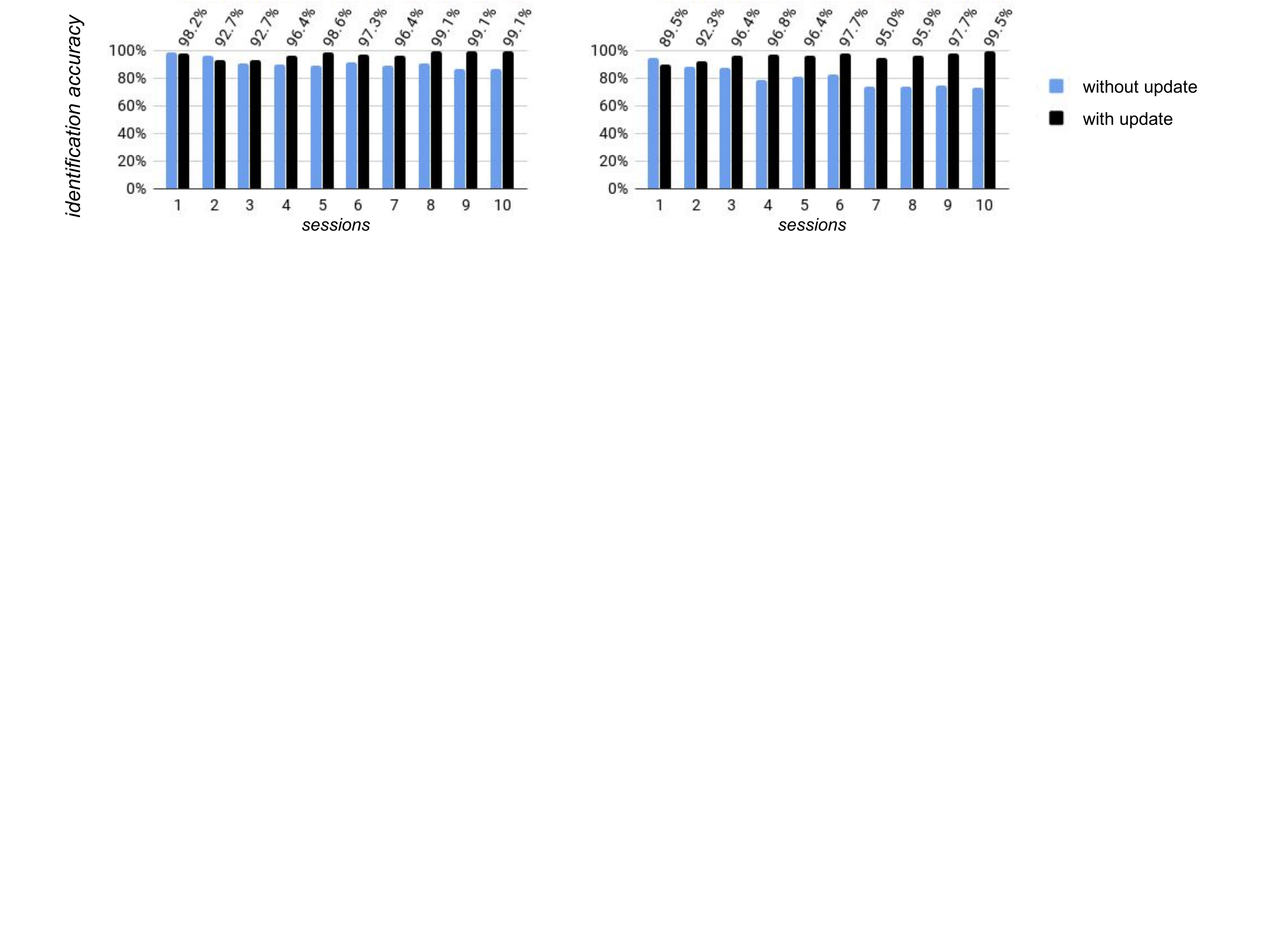}
\vspace{-4.0in}
\caption{Identification permanence performance with the glove device (left) and the camera device (right). Annotation shows results with update.}
\label{fig:ident_permanence}
\end{figure*}

\subsection{Permanence Experiments and Results}

We use the third dataset to study the long term performance of our prototype system by running the authentication procedure and the identification procedure for the five signals in each session. The change of authentication acceptance rate is shown in Fig. \ref{fig:auth_permanence} (with the same setup as that in the authentication experiments described above, and a decision threshold at $FAR = 10^{-4}$). The figure shows slight performance degradation with time. We can update the template at the end of the session to prevent such a degradation for future sessions. The new template is updated as an affine combination of the old template and the new signal, \textit{i.e.,} $\hat{\mathbf{t}}_i \leftarrow (1 - \lambda) \hat{\mathbf{t}}_i + \lambda \mathbf{s}_i$, where $\lambda$ is the update factor set to 0.1 in this experiment. Additionally, we can also both update the template and retrain the SVM classifers with both the old signals and new signals, which can further maintain or even improve performance. 

The change of identification accuracy is shown in Fig. \ref{fig:ident_permanence} (with the same setup as that in the identification experiments described above). The figure shows slight performance degradation with time. Similarly, the accuracy can be improved significantly if the CNN and the SVM classifiers are both retrained with the new signals at the end of the session. We believe that for some users merely 5 signals at registration cannot fully capture the uncertainty of the writing behavior, even with our data augmentation methods. In practice, typing a password can always be employed as a fallback. On the smartphone, if the user does not unlock it immediately with fingerprint or face, the smartphone will ask the user to type a password. If the password passes, it will update the fingerprint or face template accordingly. Such a strategy can also be utilized in our framework since showing a virtual keyboard to type a password can always be an backup option, though it is inconvenient.

\subsection{Comparison to Existing Works}

A comparison to existing works which also use in-air-handwriting is shown in the Table \ref{tb:com}. The major characteristics that differentiate our framework from them are as follows. First we use a data driven method by designing features and utilizing machine learning models, instead of crafting algorithms calculating a matching distance for authentication. Second, we avoid exhaustively searching and comparing the whole account database for user identification. Third, we evaluate performance of our framework under active spoofing attacks and with a time span of near a month, which is usually omitted by existing works. Fourth, our system has a significant performance improvement on a dataset with reasonable size. A more complete list and comparison of related work in this area is provided in Appendix C.

\subsection{Comparison to Password and Other Biometrics}

In Table \ref{tb:compb} we present a comparison of FMCode with password and biometrics based system. The results shown here are obtained from different publications with various datasets, which merely show limited information as an intuition about their performance, instead of the performance with serious and strongly supervised evaluation. First we show the classification accuracy in terms of EER. Here FMCode is comparable to fingerprint (on FVC2004 \cite{Fingerprint} among all the datasets), face, iris, and signature. Comparing to biometrics, a considerable portion of discrimination capability comes from the large content space of the in-air-handwriting. Next we show the equivalent key space in number of bits. For password used in device unlock, the commonly used 4-digit password (default setting on most smartphone) are considered, and for biometrics, the equivalent key space is defined by $log_2(1/FAR)$ \cite{Compare}. For FMCode we calculate key space with the corresponding FAR by setting the decision threshold at a place where the true user has 95\% successful login rate (\textit{i.e.,} 5\% FRR). The results show that FMCode is also comparable to password based login and authentication system. Due to the limited amount of data, we cannot achieve an FAR resolution lower than $5 \times 10^{-6}$, \textit{i.e.,} more than 17.6 bit key space. For the glove device, at the 5\% FRR decision threshold, the FAR is already 0 but we can only conclude that the equivalent key space is larger than 17.6 bits. In practice, recommended password key space is between 28 to 128 bits \cite{RFC4086} while web users typically choose passwords with 20 to 90 bits key space and on average it is 40 to 41 bits \cite{PasswordHabit}. However such large key space is underutilized because it is well known that people are not good at creating and memorizing strong password, and the actually entropy is much less than the whole key space \cite{PasswordEntropy}. Moreover, since password must contain letters that can be typed on keyboard, efficient password guessing strategies such as dictionary attack further weaken the calculated password quality in number of bits. A more detailed analysis on the comparison of usability, deployability, and security with password and biometrics is provided in Appendix B.

\begin{table*}[]
\centering
\caption{Comparison to existing works.}
\label{tb:com}
\begin{tabular}{|l|c|c|c|c|c|c|}
\hline
Ref.                                    & \begin{tabular}[c]{@{}c@{}}dataset\\ size\end{tabular} & \begin{tabular}[c]{@{}c@{}}EER\\ (w/o spoof)\end{tabular} & \begin{tabular}[c]{@{}c@{}}EER\\ (w/ spoof)\end{tabular} & \begin{tabular}[c]{@{}c@{}}Identification\\ Accuracy\end{tabular} & Device     & Algorithm       \\ \hline
FMCode (glove)                                 & 105 (210)                                               & 0.1\%                                                     & 2.2\%                                                    & 96.7\% (99.9\%)                                                            & data glove & SVM / CNN       \\ \hline
FMCode (camera)                                 & 109 (218)                                               & 0.5\%                                                     & 4.4\%                                                    & 94.3\% (99.5\%)                                                           & Leap Motion & SVM / CNN       \\ \hline
Liu et al.\cite{uWave}                & 20 $\sim$ 25                                           & $\sim$ 3\%                                                & $\sim$10\%                                               & 88 $\sim$ 98.4\%                                                  & Wii remote & DTW             \\ \hline
Bailador et al.\cite{Madrid-Analysis} & 96                                                     & 1.8\% $\sim$ 2.1\%                                        & $\sim$ 5\%                                               & N/A                                                               & smartphone & DTW, Bayes, HMM \\ \hline
Bashir et al.\cite{SmartPen-RDTW}     & 40                                                     & $\sim$1.8\%                                               & N/A                                                      & 98.5\%                                                            & custom digital pen & DTW             \\ \hline
Chan et al.\cite{LeapHand}             & 16                                                     & 0.8\%                                                     & N/A                                                      & 98\%                                                              & Leap Motion & random forest   \\ \hline
Tian et al.\cite{KinWrite}            & 18                                                     & $\sim$2\%                                                 & N/A                                                      & N/A                                                               & Kinect     & DTW             \\ \hline
\end{tabular}
\end{table*}

\begin{table*}[]
\centering
\caption{Comparison to password and biometrics.}
\label{tb:compb}
\begin{tabular}{lcccccclcl}
metric                         & \begin{tabular}[c]{@{}c@{}}FMCode\\ (glove)\end{tabular} & \begin{tabular}[c]{@{}c@{}}FMCode\\ (camera)\end{tabular} & \begin{tabular}[c]{@{}c@{}}Password\\ (online login)\end{tabular} & \begin{tabular}[c]{@{}c@{}}Password\\ (device unlock)\end{tabular} & Fingerprint       & Face               & \multicolumn{1}{c}{Iris}   & \multicolumn{1}{c}{Signature} &                                               \\ \cline{1-9}
EER (w/o spoof)                & 0.1\%                                                    & 0.5\%                                                     & N/A                                                               & N/A                                                                & 0.28\%$\sim$2\% \cite{Fingerprint} & 2.6\%$\sim$8.6\% \cite{parkhi2015deep} & 0.11\% \cite{Compare} & 1\%$\sim$3\% \cite{Signature}                & \begin{tabular}[c]{@{}l@{}}\\ \vphantom{a}\end{tabular} \\
Key Space (bits) & $>$17.6                                                     & 16                                                        & 20$\sim$90 \cite{PasswordHabit}                                          & 13.3                                                               & 13.3 \cite{Compare}              & N/A                & 19.9 \cite{Compare}   & N/A                           &                                              
\end{tabular}
\end{table*}

\section{User Evaluation}

\begin{figure}[]
\begin{center}
\includegraphics[width=3.2in]{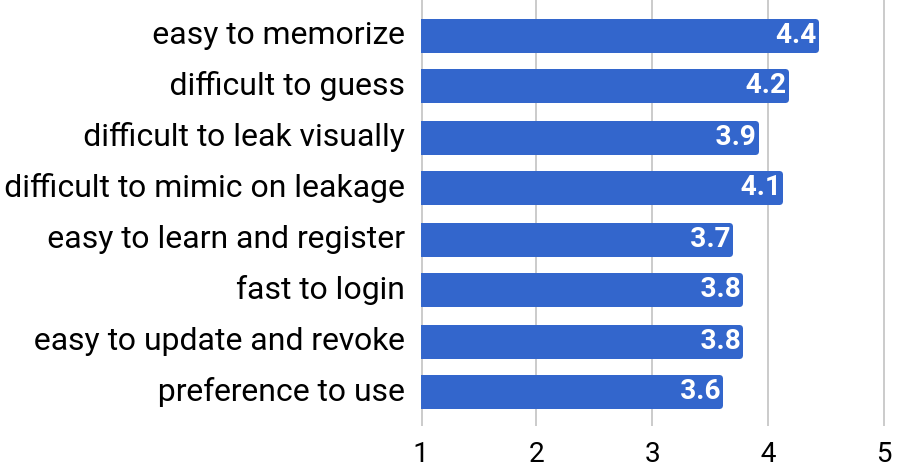}
\end{center}
   \caption{User evaluation scores.}
\label{fig:usability}
\end{figure}

We investigated the usability of our framework by taking questionnaire from 30 users with the experience of both the glove device and the camera device. First, the users evaluate various aspects of our in-air-handwriting based login framework with a score from 1 (strongly disagree) to 5 (strongly agree). The results are shown in Fig. \ref{fig:usability}.

% \begin{enumerate}
%     \item Is it easy to memorize?
%     \item Is it difficult to guess?
%     \item Is it difficult to leak visually?
%     \item Is it difficult to mimic on leakage?
%     \item Is it easy to learn and register?
%     \item Is it fast to login?
%     \item Is it easy to update and revoke?
%     \item Will you prefer to use it as the primary login method?
% \end{enumerate}

\begin{figure}[]
\begin{center}
\includegraphics[width=3.5in]{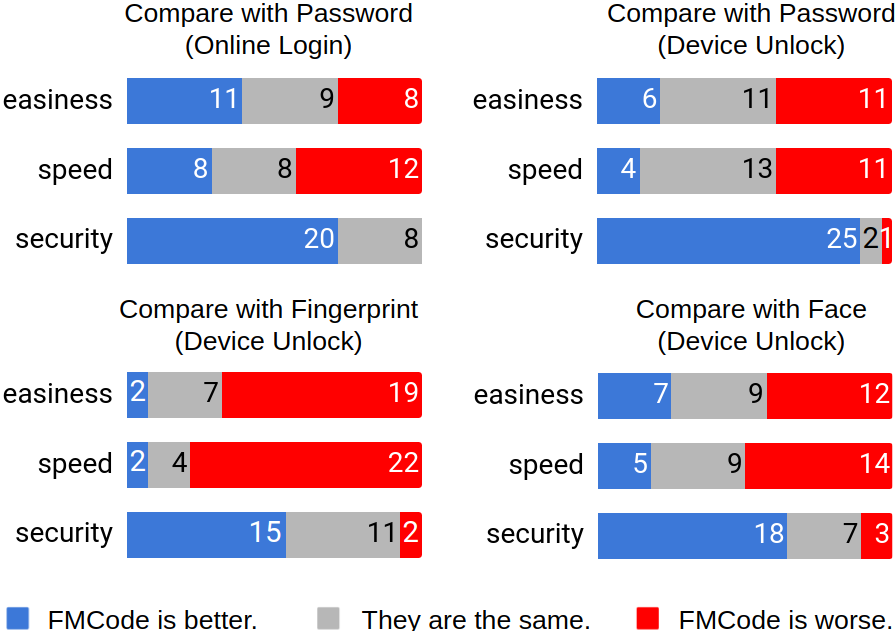}
\end{center}
   \caption{User evaluation with comparison.}
\label{fig:user_compare}
\end{figure}

Second, we ask the user to compare our framework with the password based systems and biometrics including fingerprint and face, on the easiness of usage, login speed and security. The user has three options: (a) our framework is better, (b) they are the same or difficult to decide which one is better or worse, (c) our framework is worse. The results are shown in Fig. \ref{fig:user_compare}. We can see that the user has a mixed attitude on the usability compared to traditional password, but clearly FMCode can not compete biometrics like fingerprint and face in usability. However, the majority of the users feel that FMCode is more secure than traditional password, and more than half of them feel it is more secure than fingerprint and face.

Third, we ask the following questions:

1) Compared to password, our framework fuses handwriting convention. Is this characteristic important?

2) Compared to biometrics, our framework allows change and revoke the in-air-handwriting passcode, which is unlinked to personal identity. Is this characteristic important?

Among the surveyed users, 89\% and 82\% of them answer ``important'' for the first and second characteristics respectively. Combined with the previous results, we can conclude that FMCode does not intend to replace existing password-based solution or biometrics. Instead, due to its unique characteristics that passwords and biometrics lack, FMCode is suitable in scenarios where such characteristics matter and where passwords and biometrics are not applicable, for example, login over gesture interface on VR headset or in operating theater.

At last, we ask the user which type of device is preferred between a wearable device and a contactless device for hand motion tracking. 21\% of the users choose the wearable device and the other 79\% choose the contactless device.

\section{Discussion}

\subsection{Other Attacks}

There are other potential attacks on our authentication system if some of our security assumptions do not hold. For example, the attacker may be able to access the server's storage to steal the templates. Due to the inherent fuzziness in the authentication step, traditional hash is not able to protect the templates like the hashed password. One possible solution that we are working on is to adapt the deep CNN to generate a fuzzy hash and further a set of keys for each account using the signals at registration. The templates can be encrypted by each of the key to form a pool. At login time, the same neural network is applied on the signal in the login request to obtain a fuzzy hash and a key. It is highly possible that this key can decrypt an encrypted templates in the pool, if the signal is generated by the legitimate user performing the correct FMCode. After the template decryption, the authentication procedure can continue. If the templates can be successfully decrypted by the generated key from the signal in the login request, we can not conclude that the user is authenticated. As we have shown in the identification results, the ID is easier to spoof without a verification step, thus an impostor with the knowledge of the ID and the passcode may also be able to derive an acceptable key. However, the attacker that steals the templates does not have such knowledge.

Another possible threat is man-in-the-middle attacks, where the attacker intercepts and records the messages between the server and the client. We believe this is not critical because machine-to-machine authentication and message encryption can be used first to allow the client and the server to communicate securely. Existing technologies such as SSL/TLS and public key infrastructure (PKI) over the Internet can be leveraged to build a secure communication channel before user login, similar to the case of password-based login on most websites over the Internet nowadays.

On the user side, record and replay attacks must also be handled. The authentication request require to present a multi-dimensional signal about the physical states of the hand during the writing. In a preliminary experiment, we found that it is difficult to synthesis such a signal from a video recorded by monolens camera or depth camera placed 1 meter away from the user. The main reason is the drastic spatial resolution drop with distance, which cause limited hand skeleton estimation accuracy. A seconary reason is the posture uncertainty of the user and the motion blur due to insufficient frame rate of the camera. If the camera is close to the user, the user might get alerted, just like if someone within a proximity is watching a user typing his or her password, the user will be alerted and stop typing. More details about this type of attack will be presented in future work.

\section{Related Works}

Traditional user authentication methods can be based on physiological traits (\textit{e.g.,} biometrics, like fingerprint and iris), knowledge (\textit{e.g.,} password), and possession (\textit{e.g.,} a card or a token). Unfortunately, none of them are perfect because physiological traits can be copied (\textit{e.g.} iris spoofing by taking eye pictures), knowledge can be forgotten (\textit{e.g.} forgotten password), and possession can be lost or faked (\textit{e.g.} stolen or copied card). Currently, password and biometrics are the most widely adopted authentication approaches, but they both have shortcomings. On the password side, remembering more and more passwords is a problem for the user, because a memorable password is usually not strong, while a strong password is difficult to remember \cite{PasswordHabit}. Most users just reuse the same password for many sites \cite{PasswordReuse}, and they rarely update it or only update it in certain predictable patterns \cite{PasswordUpdate}. As a result, password based system works far from optimal and there are numerous works which have made attempts to replace password \cite{Compare, Quest}. On the biometrics side, although identity verification using fingerprint \cite{Fingerprint} and iris \cite{Iris} is convenient, inexpensive and more secure than password in some cases, biometrics are inherently not secret, and they are irrevocable after leakage. Moreover, biometric authentication is directly linked to a person and hence sensitive to privacy issues. FMCode fuses a secret with the handwriting behavior, which has the advantages of both password and biometrics. Besides, FMCode is able to defend spoofing attack, while such attacks are surprisingly effective for biometrics \cite{FingerprintSpoofing}.

Early gesture and body motion based authentication methods use the accelerometer to track simple motions like shaking \cite{GIT, ArmSweep, ShakeWell}, walking \cite{ArmSwing, AreYouWithMe}, or other predefined gesture patterns \cite{Feasibility}, which proves the feasibility of such methods. In-air-handwriting captured by a smartphone\cite{Smartphone, Greece, Madrid-Analysis}, handhold remote control devices \cite{uWave} or a smartwatch \cite{HandWorn} has better potentials because the complex motions contain more information unique to each user. However, the user behavior of writing with such devices different from pen is unnatural and inconsistent. In-air-handwriting with fingertip is also investigated with camera based gesture interface such as Kinect \cite{KinWrite, Kinect-DTW, WaveToMe} and Leap Motion \cite{LeapPassword, CleanRoom, LeapHand, Poland}. The major drawback of such methods is strong constraints on the size, speed, and orientation of a user's movement, because these cameras are set up at a fixed place. These constraints have a negative influence on both usability and authentication performance, as can be shown by the performance difference of our framework with the two different types of devices. Wearable monolens cameras such as Google Glass \cite{VSig} would have certain advantages. However the vision understanding and pose estimation of hands in changing environment is difficult. For gesture based authentication, the critical technical challenge is developing good features and classifiers that can tolerate variations and long term behavior changes of the captured motions of the same user as well as distinguishing different users with high accuracy. For gesture based identification, the critical technical challenge is designing efficient way to index each account by the gesture signal or features for fast search.

Handwriting and signature has been studied extensively in the past, but FMCode is fundamentally different from handwriting \cite{Airwriting, OnlineWriting} or hand gesture recognition \cite{AirDraw, Serendipity, Tomo}, \textit{i.e.,} the former identifies and verifies the writer while the latter recognizes the content of writing or meaning of the gesture regardless of the writer. Since a recognition system of handwriting or hand gesture tries to group similar motion patterns to the same class, if they are used as login systems, the recognized content is essentially another form of a password, which is vulnerable when the content of the writing is leaked. Besides, our framework is different from online signature verification \cite{Signature}. We allow the user to draw arbitrary patterns instead of signatures (\textit{i.e.,} names), and we capture 3D motions of fingertip instead of movement of a pen or digital pen with sensors \cite{SmartPen-RDTW, DigiPen} on a 2D surface. Our method targets the application scenarios with gesture interface but lacking tradition input interfaces like keyboard or mouse. For the same reason, we omit comparison to existing 2D free gesture based system on touchscreen \cite{frank2013touchalytics, gong2016forgery, yang2016free}.

\section{Conclusions}

In this paper, we present FMCode, a user authentication and identification framework for applications using gesture input interface. FMCode does not completely replace password or biometrics in all use cases, but it can be an alternative way for user login over gesture interface where password and passive biometrics are inconvenient or not applicable. Our prototype system and experimental results show great potential of this method. However, FMCode also has limitations under certain security assumptions, which require further investigation, including methods of direct key generation from in-air-handwriting signals and performance under record and replay attack.

\newpage

\bibliographystyle{IEEEtran}
\bibliography{reference}

\section*{Appendix}

\subsection{Cost of Computing and Storage}

We implemented our prototype system in Python with sklearn library and TensorFlow \cite{TensorFlow}. Preprocessing a single signal cost about 25 ms for the glove device (excluding the state estimation step since it is running on the device instead of the client machine) and 100 ms for the camera device, where filtering is the bottleneck. In authentication, generate template cost 2.1 ms and training the SVM cost 30 ms for each account; classification of each signal cost less than 1 ms, which is negligible compared to the time for writing the string. The time consumption measured here does not contain loading the data and model from disk to the main memory, because the disk speed varies significantly due to the page cache in the main memory and our dataset is small enough to be fully fit in the cache, which is quite different when used in real world scenarios. This measurement is conducted with a single threaded program on a workstation with Intel Xeon E3-1230 CPU (quad-core 3.2GHz, 8M cache) and 32 GB main memory. For identification, training the deep CNN for 20 epochs requires around 7 minutes using only CPU, while with a powerful GPU (Nvida GTX 1080 Ti in our case) this can drastically decrease to around 30 seconds; classification using the CNN costs less than 1 ms with only CPU. The space cost for template storage and the amount of data needed to be transferred between the server and the client is proportional to the signal length. If each sensor axis of a sample is represented in single precision floating point number (four bytes), the average space cost of a signal is 8.6 KB with our datasets for both devices (they both output data with 9 sensor axis). If all parameters are represented in single precision floating point number, storing the SVM classifiers costs 79 KB per account on average, and storing the CNN itself requires around 4 MB because of the 1 million weights and biases parameters.

\subsection{Comparison to Password and Biometrics in Usability, Deployability, Security}

\begin{table*}[]
\small
\centering
\caption{Usability, Deployability, and Security Evaluation of FMCode.}
\label{table:uds}
\begin{tabular}{|l|l|l|l|l|l|}
\hline
Usability             & FMCode   & Deployability                      & FMCode & Security                            & FMCode   \\ \hline
Memory effortless     & Maybe (+) & Accessible                         & Yes    & Resilient to physical observation   & Yes (+)   \\ \hline
Scalable for users    & Maybe (+) & Scalable                           & Yes    & Resilient to targeted impersonation & Maybe (+) \\ \hline
Nothing to carry      & Yes      & Server compatible                  & Maybe (-)   & Resilient to throttled guessing     & Maybe    \\ \hline
Physically effortless & No       & Browser compatible                 & Maybe (-)    & Resilient to unthrottled guessing   & No       \\ \hline
Easy to learn         & Yes      & Mature                             & No (-) & Resilient to theft                  & Yes (+)  \\ \hline
Efficient to use      & Yes      & Non-proprietary                    & Yes    & No trusted third party required     & Yes      \\ \hline
Infrequent errors     & Maybe (-)    & Configurable\cite{Engineering}       & Yes    & Requiring explicit consent          & Yes      \\ \hline
Easy to recovery      & Yes      & Developer friendly\cite{Engineering} & Yes    & Unlinkable                          & Yes      \\ \hline
\end{tabular}
\end{table*}

\begin{table*}[]
\small
\centering
\caption{Comparison of experimental results and algorithms of related works.}
\label{table:related_works}
\begin{tabular}{|l|l|l|l|l|l|l|l|}
\hline
Related Works                           & \begin{tabular}[c]{@{}l@{}}\# of \\ subjects\end{tabular} & Device                                                                                  & \begin{tabular}[c]{@{}l@{}}Experiment \\ Timespan\end{tabular} & \begin{tabular}[c]{@{}l@{}} EER\end{tabular}     & \begin{tabular}[c]{@{}l@{}}Claimed \\ Accuracy\end{tabular}  & \begin{tabular}[c]{@{}l@{}}Motion \\ Gesture\end{tabular}          & Algorithm                                                                        \\ \hline
Patel et al.\cite{GIT}                & NA                                                        & \begin{tabular}[c]{@{}l@{}}Cellphone w/ \\ accelerometer \end{tabular} & NA                                                             & NA            & NA                                                           & shake                                                                 & \begin{tabular}[c]{@{}l@{}}static \\ threshold\end{tabular}                      \\ \hline
Okumura et al.\cite{ArmSweep}         & 12 $\sim$ 22                                              & \begin{tabular}[c]{@{}l@{}}Cellphone w/ \\ accelerometer\end{tabular}                   & 6 weeks                                                        & 4\%                                                                   & NA                                                           & shake                                                                 & DTW                                                                              \\ \hline
Mayrhofer et al.\cite{ShakeWell}      & 51                                                        & \begin{tabular}[c]{@{}l@{}}Custom device \\ w/ accelerometer\end{tabular}    & NA                                                             & \begin{tabular}[c]{@{}l@{}}$\sim$2\%,\\ $\sim$10\%\end{tabular}       & NA                                                           & shake                                                                 & \begin{tabular}[c]{@{}l@{}}frequency \\ coherence\end{tabular}                   \\ \hline
Farella et al.\cite{Feasibility}      & 5 $\sim$ 10                                               & \begin{tabular}[c]{@{}l@{}}PDA w/ \\ accelerometer\end{tabular}                         & NA                                                             & NA                                                                    & \begin{tabular}[c]{@{}l@{}}63\% \\ $\sim$ 97\%\end{tabular}  & \begin{tabular}[c]{@{}l@{}}4 specified \\ gestures\end{tabular}       & \begin{tabular}[c]{@{}l@{}}PCA / LLE \\ + kNN\end{tabular}                       \\ \hline
Lester et al.\cite{AreYouWithMe}      & 6                                                         & \begin{tabular}[c]{@{}l@{}}Custom device \\ w/ accelerometer\end{tabular}               & NA                                                             & NA                                                                    & 100\%                                                        & walk                                                                  & \begin{tabular}[c]{@{}l@{}}frequency \\ coherence\end{tabular}                   \\ \hline
Gafurov et al.\cite{ArmSwing}         & 30                                                        & \begin{tabular}[c]{@{}l@{}}Custom device \\ w/ accelerometer\end{tabular}               & NA                                                             & 10\%                                                                  & NA                                                           & walk                                                                  & \begin{tabular}[c]{@{}l@{}}frequency \\ similarity\end{tabular}                  \\ \hline
Liu et al.\cite{uWave}                & 20 $\sim$ 25                                              & \begin{tabular}[c]{@{}l@{}}Nintendo Wii \\ remote\end{tabular}                          & 1 $\sim$ 4 weeks                                               & \begin{tabular}[c]{@{}l@{}}$\sim$3\%,\\ \textgreater10\%\end{tabular} & \begin{tabular}[c]{@{}l@{}}88\% \\ $\sim$ 99\%\end{tabular}  & free writing                                                          & DTW                                                                              \\ \hline
Zaharis et al.\cite{Greece}           & 4                                                         & \begin{tabular}[c]{@{}l@{}}Nintendo Wii \\ remote\end{tabular}                          & 3 weeks                                                        & NA                                                                    & 98.20\%                                                      & free writing                                                          & \begin{tabular}[c]{@{}l@{}}statistical \\ feature matching\end{tabular}       \\ \hline
Casanova et al.\cite{Madrid-Analysis} & 96                                                        & Smartphone                                                                              & 20 days                                                        & $\sim$2.5\%                                                           & NA                                                           & free writing                                                          & \begin{tabular}[c]{@{}l@{}}DTW, Bayes, \\ HMM\end{tabular}                       \\ \hline
Lee et al.\cite{Korea}                & 15                                                        & Smartphone                                                                              & NA                                                             & NA                                                                    & 88.40\%                                                      & tap, flip, etc.                                                       & decision tree                                                                    \\ \hline
Bashir et al.\cite{SmartPen-RDTW}     & 10 $\sim$ 40                                              & Custom pen                                                                              & NA                                                             & $\sim$1.8\%                                                           & 98.50\%                                                      & \begin{tabular}[c]{@{}l@{}}alphabetic or \\ free writing\end{tabular} & RDTW                                                                             \\ \hline
Renuka et al.\cite{DigiPen}           & 12                                                        & Custom pen                                                                              & NA                                                             & NA                                                                    & $\sim$ 95\%                                                  & \begin{tabular}[c]{@{}l@{}}alphabetic or \\ free writing\end{tabular} & NA                                                                               \\ \hline
Aslan, et al.\cite{CleanRoom}         & 13                                                        & Leap Motion                                                                             & NA                                                             & $\sim$10\%                                                            & NA                                                           & \begin{tabular}[c]{@{}l@{}}2 specified \\ gestures\end{tabular}       & DTW                                                                              \\ \hline
Nigam et al.\cite{LeapPassword}       & 150                                                       & Leap Motion                                                                             & NA                                                             & NA                                                                    & 81\%                                                         & free writing                                                          & \begin{tabular}[c]{@{}l@{}}statistical feature \\ classification\end{tabular} \\ \hline
Chan et al.\cite{LeapHand}            & 16                                                        & Leap Motion                                                                             & NA                                                             & 0.80\%                                                                & \textgreater 99\%                                            & free writing                                                          & random forest                                                                    \\ \hline
Piekarczyk et al.\cite{Poland}        & 4 $\sim$ 5                                                & Leap Motion                                                                             & NA                                                             & NA                                                                    & \begin{tabular}[c]{@{}l@{}}88\% \\ $\sim$ 100\%\end{tabular} & \begin{tabular}[c]{@{}l@{}}4 specified \\ gestures\end{tabular}       & \begin{tabular}[c]{@{}l@{}}DCT + DTW \\ + kNN / LSH\end{tabular}                 \\ \hline
Tian et al.\cite{KinWrite}            & 18                                                        & Kinect                                                                                  & 5 months                                                       & $\sim$2\%                                                             & NA                                                           & free writing                                                          & DTW                                                                              \\ \hline
Sajid et al.\cite{VSig}               & 10                                                        & Google Glass                                                                            & NA                                                             & NA                                                                    & 97.50\%                                                      & free writing                                                          & \begin{tabular}[c]{@{}l@{}}PCA + GMM \\ Clustering + DTW\end{tabular}         \\ \hline
Wu et al.\cite{Kinect-DTW}            & 20                                                        & Kinect                                                                                  & NA                                                             & 1.89\%                                                                & NA                                                           & 8 gestures                                                            & DTW                                                                              \\ \hline
Hayashi et al.\cite{WaveToMe}         & 36                                                        & Kinect                                                                                  & 2 weeks                                                        & \begin{tabular}[c]{@{}l@{}}0.5\% \\ $\sim$ 1.6\%\end{tabular}         & NA                                                           & hand waving                                                           & SVM                                                                              \\ \hline
\end{tabular}
\end{table*}

We evaluated the usability, deployability and security of FMCode using the criteria provided by \cite{Quest}, and the result is shown in Table \ref{table:uds}. We added two aspects in deployability, ``configurable'' and ``developer friendly'' mentioned in \cite{Engineering}. Each usability, deployability and security item is evaluated by whether our FMCode method possess the characteristics, and a plus / minus sign means that our method is better / worse than password. We explain a few items here but the readers are recommended to refer to \cite{Quest} and \cite{Engineering} for the definition of these characteristics. In general, compared with password and biometrics, FMCode achieves nearly all their usability and deployability characteristics, and it contains the security advantages from both password and biometrics.

On the usability side, nothing to carry means the user does not to present a physical item such as a card. Though our prototype system uses a glove, the glove is treated as a general gesture input device, just like password does not require the user to bring a keyboard everywhere. Arguably, FMCode has less memory burden because in section 6 we shows the discrimination capability comes from 3 layers of information instead of just the combination of characters in password, so memorable FMCode is not necessarily weak any more. However, there might be potentially more frequent rejection of legitimate user login than password because the internal fluctuation in the stability of finger motion. 

On the deployability side, since the server only needs to store the template as a secret, FMCode is similar as those behavior biometrics without using special devices (\textit{e.g.,} keyboard or mouse dynamics), thus it can be compatible with most server with small software modification. FMCode can be supported by browsers on devices with gesture input interface, where the FMCode input of a user can be treated as a very long password and sent to the server over the Internet, similar as the case of logining in a website using password.

On the security side, FMCode can withstand spoof under semantic or visual disclosure and targeted impersonation to a certain extent, which is more like biometrics than password. But unlike passive biometrics, FMCode is changable and more difficult to steal. For example, an attacker can collect a user's fingerprint from a cup after the user has touched it, but FMCode can only be recorded when the user performs it. FMCode suffers from server storage leakage and internal observer if the template is not properly protected, and such proper template protection might be difficult because of fussiness in alignment and matching. Also it shares all other security deficiencies of password and biometrics under attacks such as brute-force guessing/collision, dictionary guessing, phishing and cross-site leakage.

Compared with traditional password, FMCode uses handwriting, which makes it a behavior biometrics, and this provides certain protection under semantic or visual disclosure of the passcode. On the other side, compared with unchangeable biometrics like fingerprint, FMCode keeps most of the advantages of a password such as revocability and privacy preserving. This also allows one user to have more than one FMCode, different from traditional behavior biometrics like gait or voice in active speaker recognition. The most similar authentication techniques would be online signature verification (technically speaking, handwriting verification), or graphical password with stroke dynamics, but most of they assume a 2D pressure sensitive touchscreen instead of writing in the air.

\subsection{Comparison of Related Works}

A comparison of related works is shown in Table \ref{table:related_works}.

\end{document}